\documentclass[10pt,twocolumn,letterpaper]{article}

\usepackage{times}
\usepackage{epsfig}
\usepackage{graphicx}
\usepackage{amsmath}
\usepackage{amssymb}
\usepackage{mathtools}
\usepackage{subcaption}
\usepackage{algorithm,algorithmic}
\usepackage[T1]{fontenc}
\usepackage[hyphens]{url}
\usepackage{booktabs}
\usepackage{makecell}
\usepackage{multirow}
\usepackage{ragged2e}
\usepackage{cvpr}

\usepackage[pagebackref=true,breaklinks=true,letterpaper=true,colorlinks,bookmarks=false]{hyperref}

\cvprfinalcopy

\def\cvprPaperID{3815}

\newcommand{\INDSTATE}[1][1]{\STATE\hspace{#1\algorithmicindent}}

\ifcvprfinal\fi
\begin{document}
\title{Accelerating Convolutional Neural Networks via Activation Map Compression}

\author{Georgios Georgiadis\\
Samsung Semiconductor, Inc.\\
{\tt\small g.geor@samsung.com}
}

\maketitle
%\thispagestyle{empty}

%%%%%%%%% ABSTRACT
\begin{abstract}
   The deep learning revolution brought us an extensive array of neural network architectures that achieve state-of-the-art performance in a wide variety of Computer Vision tasks including among others, classification, detection and segmentation. In parallel, we have also been observing an unprecedented demand in computational and memory requirements, rendering the efficient use of neural networks in low-powered devices virtually unattainable. Towards this end, we propose a three-stage compression and acceleration pipeline that sparsifies, quantizes and entropy encodes activation maps of Convolutional Neural Networks. Sparsification increases the representational power of activation maps leading to both acceleration of inference and higher model accuracy. Inception-V3 and MobileNet-V1 can be accelerated by as much as $1.6\times$ with an increase in accuracy of $0.38\%$ and $0.54\%$ on the ImageNet and CIFAR-10 datasets respectively. Quantizing and entropy coding the sparser activation maps lead to higher compression over the baseline, reducing the memory cost of the network execution. Inception-V3 and MobileNet-V1 activation maps, quantized to $16$ bits, are compressed by as much as $6\times$ with an increase in accuracy of $0.36\%$ and $0.55\%$ respectively.
\end{abstract}

%%%%%%%%% BODY TEXT
\section{Introduction}
\begin{figure}[t]  
  \begin{subfigure}{\columnwidth}
    \centering
    \includegraphics[width=\columnwidth]{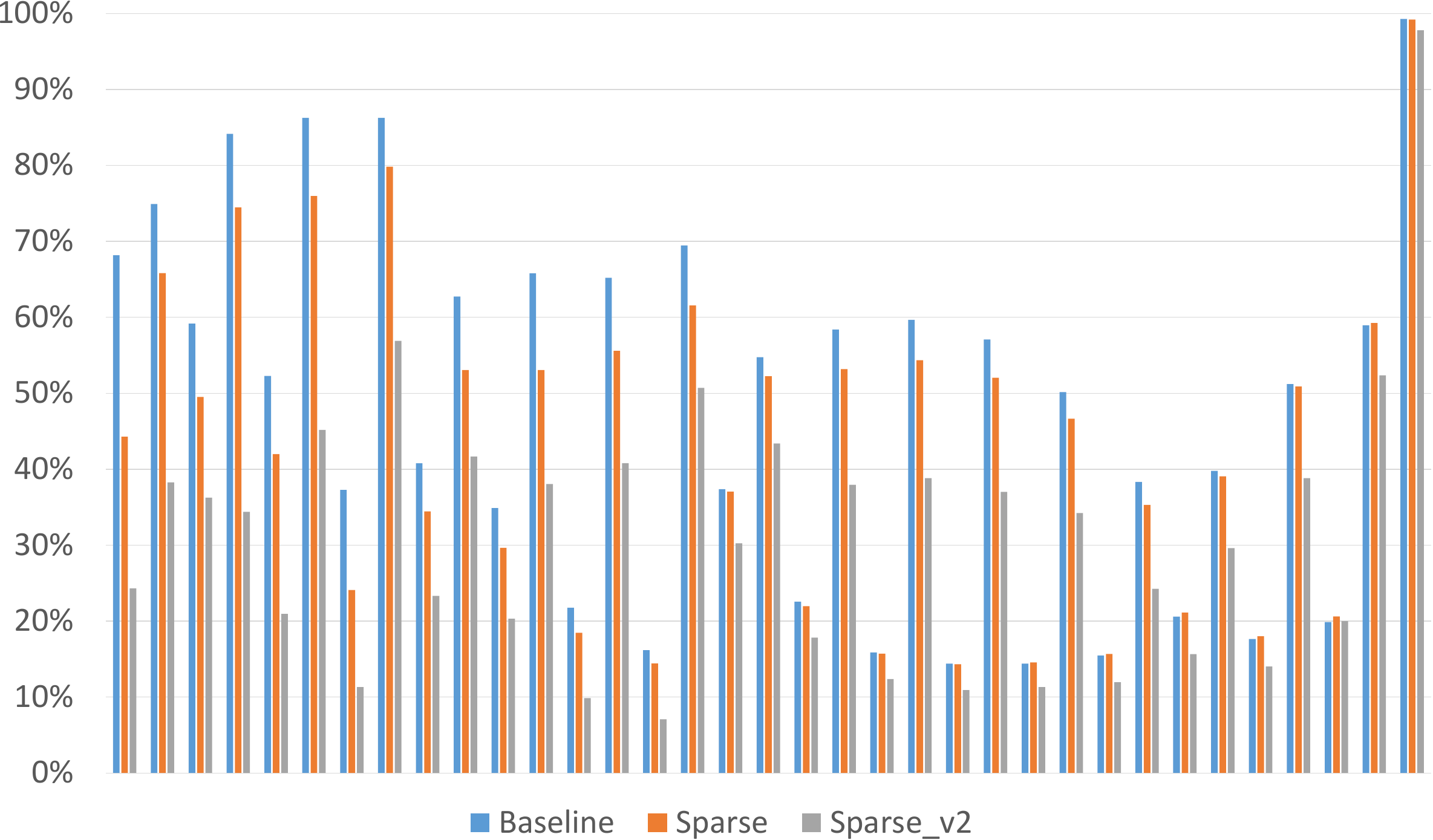}
  \end{subfigure}\\
  \begin{subfigure}{\columnwidth}
    \centering
    \includegraphics[width=0.95\columnwidth]{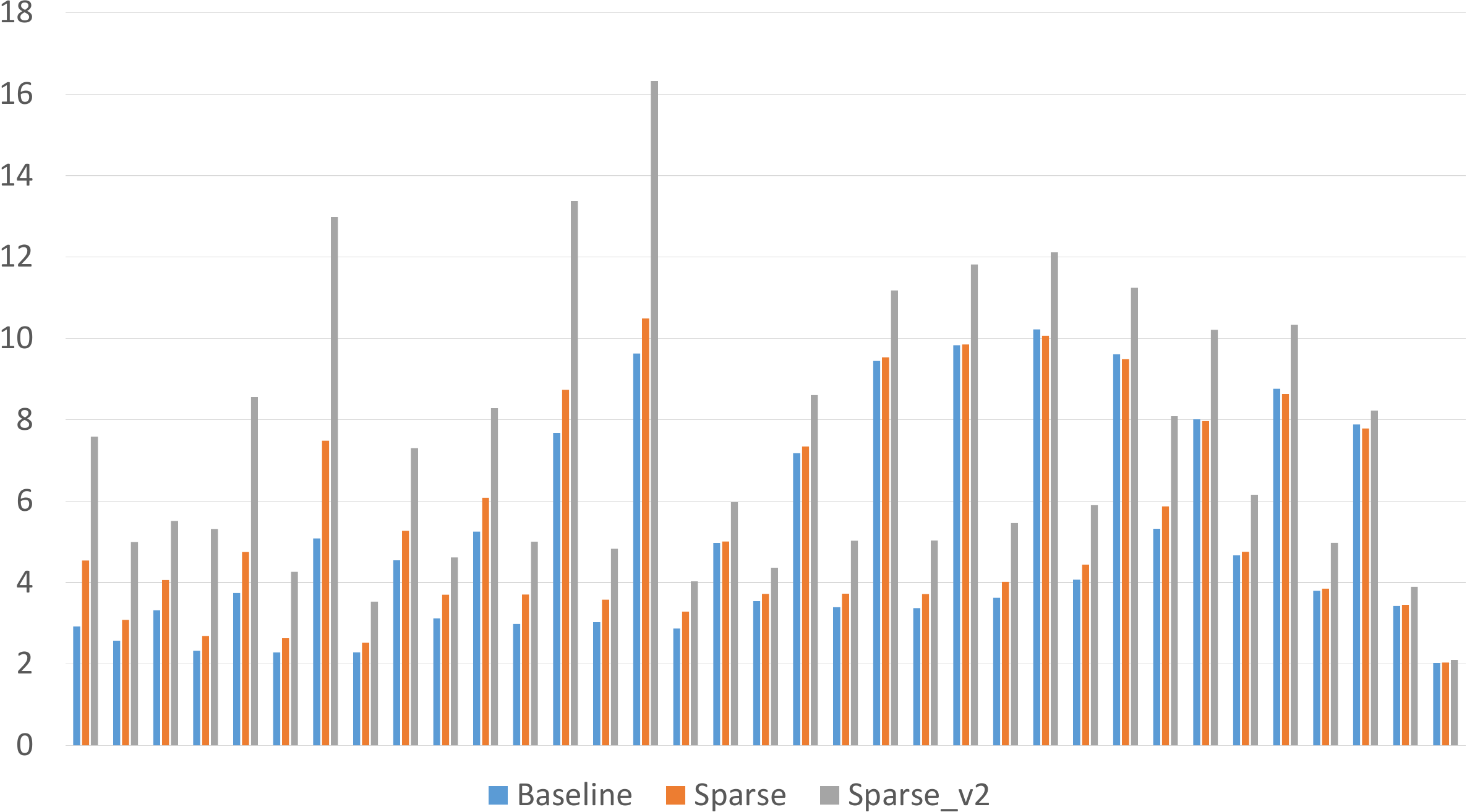}
  \end{subfigure}%  
  \caption{Percentage of non-zero activations (above) and compression gain (below) per layer of ResNet-34. Left to right order corresponds to first to last layer in the network. Baseline corresponds to the original model, while sparse and sparse\_v2 correspond to our sparsified models.}
  \label{fig:inc3-spars}
\end{figure}
With the resurgence of Deep Neural Networks occurring in 2012 \cite{krizhevsky2012}, efforts from the research community have led to a multitude of neural network architectures \cite{he2016, simonyan2014, szegedy2016b, szegedy2016} that have repeatedly demonstrated improvements in model accuracy. The price paid for this improved performance comes in terms of an increase in computational and memory cost as well as in higher power consumption. For example, AlexNet \cite{krizhevsky2012} requires 720 million multiply-accumulate (MAC) operations and features 60 million parameters, while VGG-16 \cite{simonyan2014} requires a staggering 15 billion MAC's and features 138 million parameters. While the size and number of MAC operations of these networks can be handled with modern desktop computers (mainly due to the advent of Graphical Processing Units (GPU)), low-powered devices such as mobile phones and autonomous driving vehicles do not have the resources to process inputs at an acceptable rate (especially when the application demands real-time processing). Therefore, there is a great need to develop systems and algorithms that would allow us to run these networks under a limited resource budget.

There has been a plethora of works that attempt to meet this need via a variety of different methods. The majority of the approaches \cite{han2015,ioannou2016,jaderberg2014} attempt to approximate the weights of a neural network in order to reduce the number of parameters (model compression) and/or the number of MAC's (model acceleration). For example, pruning \cite{han2015} compresses and accelerates a neural network by reducing the number of non-zero weights. Sparse tensors can be more effectively compressed leading to model compression, while multiplication with zero weights can be skipped leading to model acceleration.
  
Accelerating a model can also be achieved not only by zero-skipping of weights but also by zero-skipping input activation maps, a fact widely used in neural network hardware accelerators \cite{Albericio2016,han2016,Kim2018,parashar2017}. Most modern Convolutional Neural Networks (CNN) use the Rectified Linear Unit (ReLU) \cite{glorot2011,hahnloser2000digital,nair2010rectified} as an activation function. As a result, a large percentage of activations are zero and can be safely skipped in multiplications without any loss. However, even though the last few activation maps are typically very sparse, the first few often contain a large percentage of non-zero values (see Fig. \ref{fig:inc3-spars} for an example based on the ResNet-34 architecture \cite{he2016}). It also so happens that the first few layers process the largest sized inputs/outputs. Hence, a network would greatly benefit in terms of memory usage and model acceleration by effectively sparsifying even further these activation maps.

However, unlike weights that are trainable parameters and can be easily adapted during training, it is not obvious how one can increase the sparsity of the activations. Hence, the first contribution of this paper is to demonstrate how this can be done without any drop in accuracy. In addition, and if desired, it is also possible to use our method to trade-off accuracy for an increase in sparsity to target a variety of different hardware specifications.

Activation map size also plays an important role in power consumption. In low-powered neural network hardware accelerators (e.g. Intel Movidius Neural Compute Stick\footnote{\url{https://ai.intel.com/nervana-nnp/}}, DeepPhi DPU\footnote{\url{http://www.deephi.com/technology/}} and a plethora of others) on-chip memory is extremely limited. The accelerator is bound to access both weights and activations from the off-chip DRAM, which requires approximately 100$\times$ more power than on-chip access \cite{horowitz14}, unless the input/output activation maps and weights of the layer fit in the on-chip memory.
In such cases, activation map compression is in fact disproportionately much more important than weight compression. Consider the second convolutional layer of Inception-V3, which takes as input a $149\times149\times32$ tensor and returns one of size $147\times147\times32$, totalling 1,401,920 values. The weight tensor is of size $32 \times 32 \times 3 \times 3$ totalling just 9,216 values. The total of input/output activations is in fact 150$\times$ larger in size than the number of weights. Compressing activations effectively can reduce dramatically the amount of data transferred between on-chip and off-chip memory, decreasing in turn significantly the power consumed. Thus, our second contribution is to demonstrate an effective (lossy) compression pipeline that can lead to great reductions in the size of activation maps, while still maintaining the claim of no drop in accuracy. Following the footsteps of weight compression in \cite{han2015}, where the authors propose a three-stage pipeline consisting of pruning, quantization and Huffman coding, we adapt this pipeline to activation compression. Our pipeline also consists of three steps: sparsification, quantization and entropy coding. Sparsification aims to reduce the number of non-zero activations, quantization limits the bitwidth of values and entropy coding losslessly compresses the activations.

Finally, it is important to note that in several applications including, but not limited to, autonomous driving, medical imaging and other high risk operations, compromises in terms of model accuracy might not be tolerated. As a result, these applications demand networks to be compressed losslessly. In addition, lossless compression of activation maps can also benefit training since it can allow increasing the batch size or the size of the input. Therefore, our final contribution lies in the regime of lossless compression. We present a novel entropy coding algorithm, called sparse-exponential-Golomb, a variant of exponential-Golomb \cite{teuhola1978}, that outperforms all other tested entropy coders in compressing activation maps. Our algorithm leverages on the sparsity and other statistical properties of the maps to effectively losslessly compress them. This algorithm can stand on its own in scenarios where lossy compression is deemed unacceptable or it can be used as the last step of our three-stage lossy compression pipeline. 

\section{Related Work}

There are few works that deal with activation map compression. Gudovskiy \etal \cite{Gudovskiy2018} compress activation maps by projecting them down to binary vectors and then applying a nonlinear dimensionality reduction (NDR) technique. However, the method modifies the network structure (which in certain use cases might not be acceptable) and it has only been shown to perform slightly better over simply quantizing activation maps. Dong \etal \cite{dong2017} attempt to predict which output activations are zero to avoid computing them, which can reduce the number of MAC's performed. However, their method also modifies the network structure and in addition, it does not \emph{increase} the sparsity of the activation maps. Alwani \etal \cite{Alwani2016} reduce the memory requirement of the network by recomputing activation maps instead of storing them, which obviously comes at a computational cost. In \cite{dhillon2018}, the authors perform stochastic activation pruning for adversarial defense. However, due to their sampling strategy, their method achieves best results when ${\sim}100\%$ samples are picked, yielding \emph{no change} in sparsity.

In terms of lossless activation map compression, Rhu \etal \cite{rhu2017} examine three approaches: run-length encoding \cite{robinson1967}, zero-value compression (ZVC) and zlib compression \cite{Gailly1996}. The first two are hardware-friendly, but only achieve competitive compression when sparsity is high, while zlib cannot be used in practice due to its high computational complexity.
Lossless weight compression has appeared in the literature in the form of Huffman coding (HC) \cite{han2015} and arithmetic coding \cite{Reagen2017}. 

Recently, many lightweight architectures have appeared in the literature that attempt to strike a balance between computational complexity and model accuracy \cite{howard2017,Iandola2016,sandler2018,zhang2017}. Typical design choices include the introduction of $1\times1$ point-wise convolutions and depthwise-separable convolutions. These networks are trained from scratch and offer an alternative to state-of-the-art solutions. When such lightweight architectures do not achieve high enough accuracy, it is possible to alternatively compress and accelerate state-of-the-art networks. Pruning of weights \cite{Guo2016,han2015,hassibi1993,lebedev2016,Li2016,Luo2017,park2016faster,ullrich2017,zhou2016} and quantization of weights and activations \cite{courbariaux2014, courbariaux2015,Gong2014,gupta2015,han2015,Hubara2016,rastegari2016,vanhoucke2011,wu2016} are the standard compression techniques that are currently used. Other popular approaches include the modification of pre-trained networks by replacing the convolutional kernels with low-rank factorizations \cite{ioannou2015,jaderberg2014,sironi2015} or grouped convolutions \cite{ioannou2016}.

One could falsely consider viewing our algorithm as a method to perform activation pruning, an analog to weight pruning \cite{han2015} or structured weight pruning \cite{louizos2017b,louizos2017,neklyudov2017,wen2016}. The latter prunes weights at a coarser granularity and in doing so also affects the activation map sparsity. However, our method affects sparsity \emph{dynamically}, and not statically as in all other methods, since it does not \emph{permanently} remove any activations. Instead, it encourages a smaller percentage of activations to fire for any given input, while still allowing the network to fully utilize its capacity, if needed. 

Finally, activation map regularization has appeared in the literature in various forms such as dropout \cite{srivastava2014dropout}, batch normalization \cite{ioffe2015batch}, layer normalization \cite{ba2016layer} and $\boldsymbol{L}_2$ regularization \cite{merity2017revisiting}. In addition, increasing the sparsity of activations has been explored in sparse autoencoders \cite{ngsparseencoders} using Kullback-Leibler (KL) divergence and in CNN's using ReLU \cite{glorot2011}. In the seminal work of Glorot \etal \cite{glorot2011}, the authors use ReLU as an activation function to induce sparsity on the activation maps and briefly discuss the use of $\boldsymbol{L}_1$ regularization to enhance it. However, the utility of the regularizer is not fully explored. In this work, we expand on this idea and apply it on CNN's for model acceleration and activation map compression.

\section{Learning Sparser Activation Maps}
\label{sec:learn_sparse}

A typical cost function, $E_0(w)$, employed in CNN models is given by:
\begin{equation}
  \label{eq:cost1}
  E_0(w) = \frac{1}{N} \sum_{n=1}^N c_n(w) + \lambda_w r ( w ),
\end{equation}
where $n$ denotes the index of the training example, $N$ is the mini-batch size, $\lambda_w \geq 0$, $w \in \mathbb{R}^d$ denotes the network weights, $c_n(w)$ is the data term (usually the cross-entropy) and $r(w)$ is the regularization term (usually the $\boldsymbol{L}_2$ norm).

\begin{figure}[t]
  \includegraphics[trim={0 0.1cm 0 0},clip,width=\linewidth]{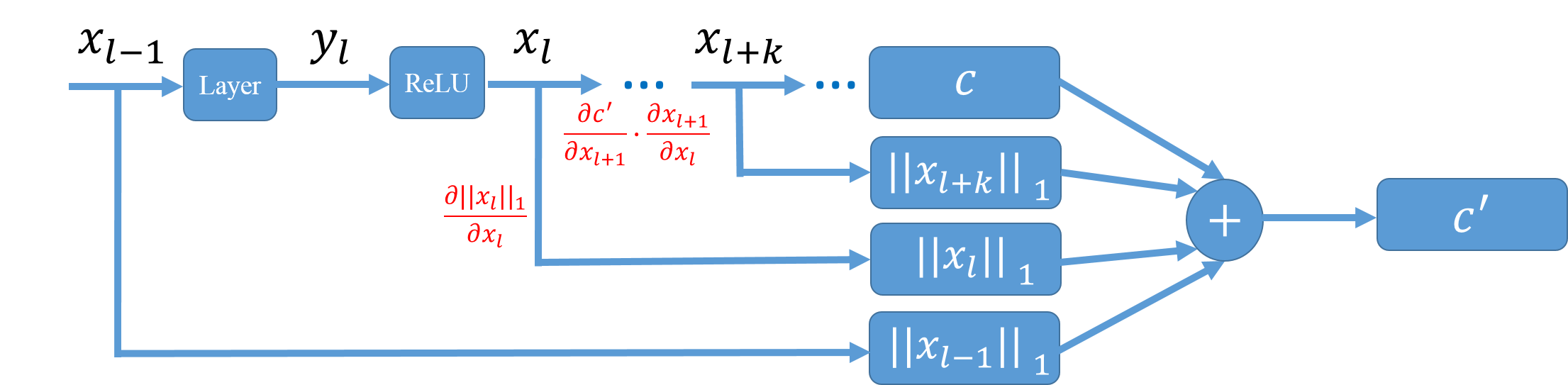}
  \caption{Computational graph based on Eq. \ref{eq:cost3}. In red we illustrate the two gradient contributions for $x_l$.}
  \label{fig:layerl}
\end{figure}

Post-activation maps of training example $n$ and layer $l \in \{1,\hdots,L\} $ are denoted by $x_{l,n} \in \mathbb{R}^{H_l \times W_l \times C_l}$, where $H_l, W_l, C_l$ denote the number of rows, columns and channels of $x_{l,n}$. When the context allows, we write $x_l$ rather than $x_{l,n}$ to reduce clutter. $x_0$ corresponds to the input of the neural network. Pre-activation maps are denoted by $y_{l,n}$. Note that since ReLU can be computed in-place, in practical applications, $y_{l,n}$ is often only an intermediate result. Therefore, we target to compress $x_{l}$ rather than $y_{l}$. See Fig. \ref{fig:layerl} for an explanatory illustration of these quantities. 

\begin{figure*}[t]
  \begin{subfigure}{.19\textwidth}
    \centering
    \includegraphics[width=.9\linewidth]{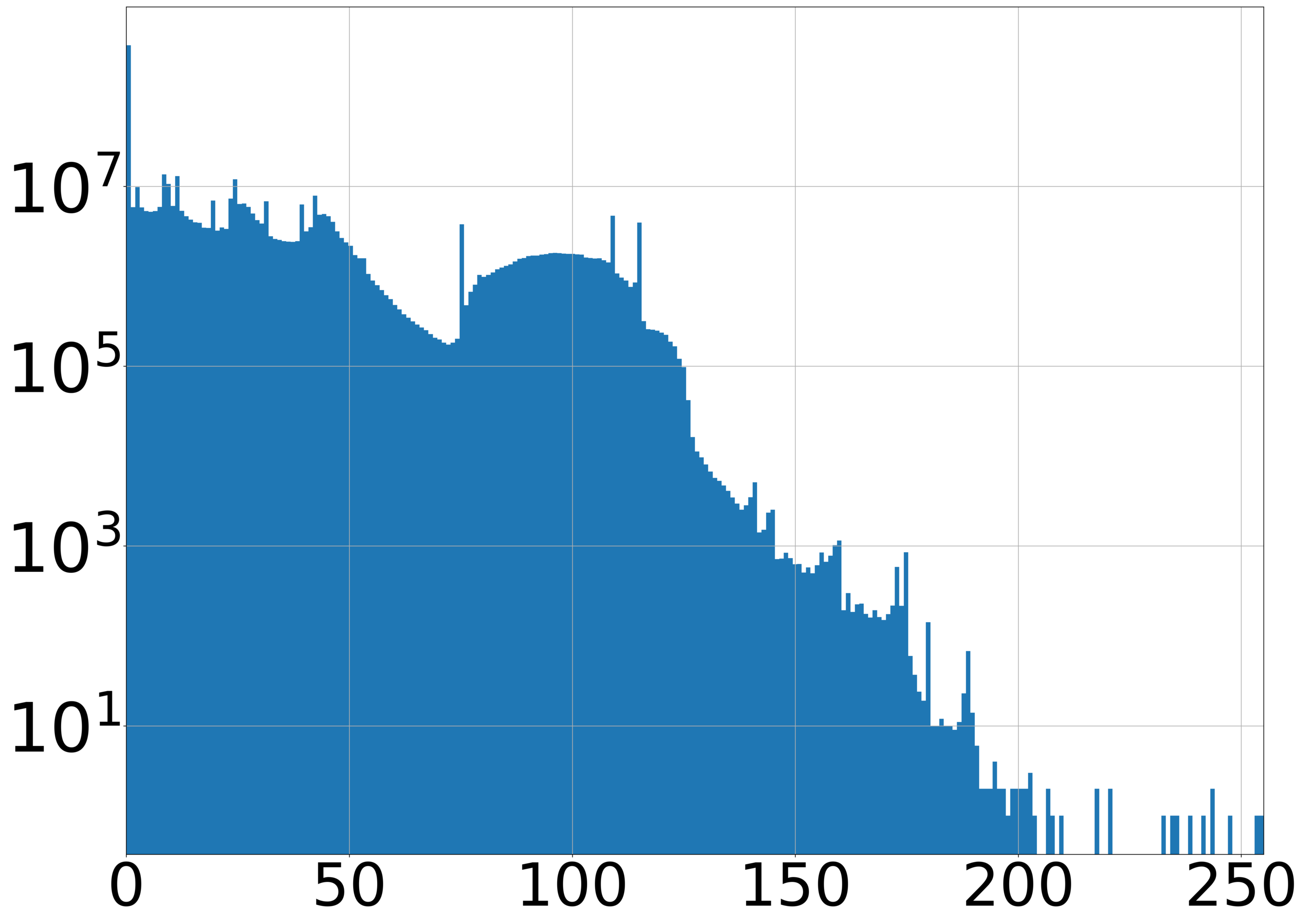}
  \end{subfigure}%
  \begin{subfigure}{.19\textwidth}
    \centering
    \includegraphics[width=.9\linewidth]{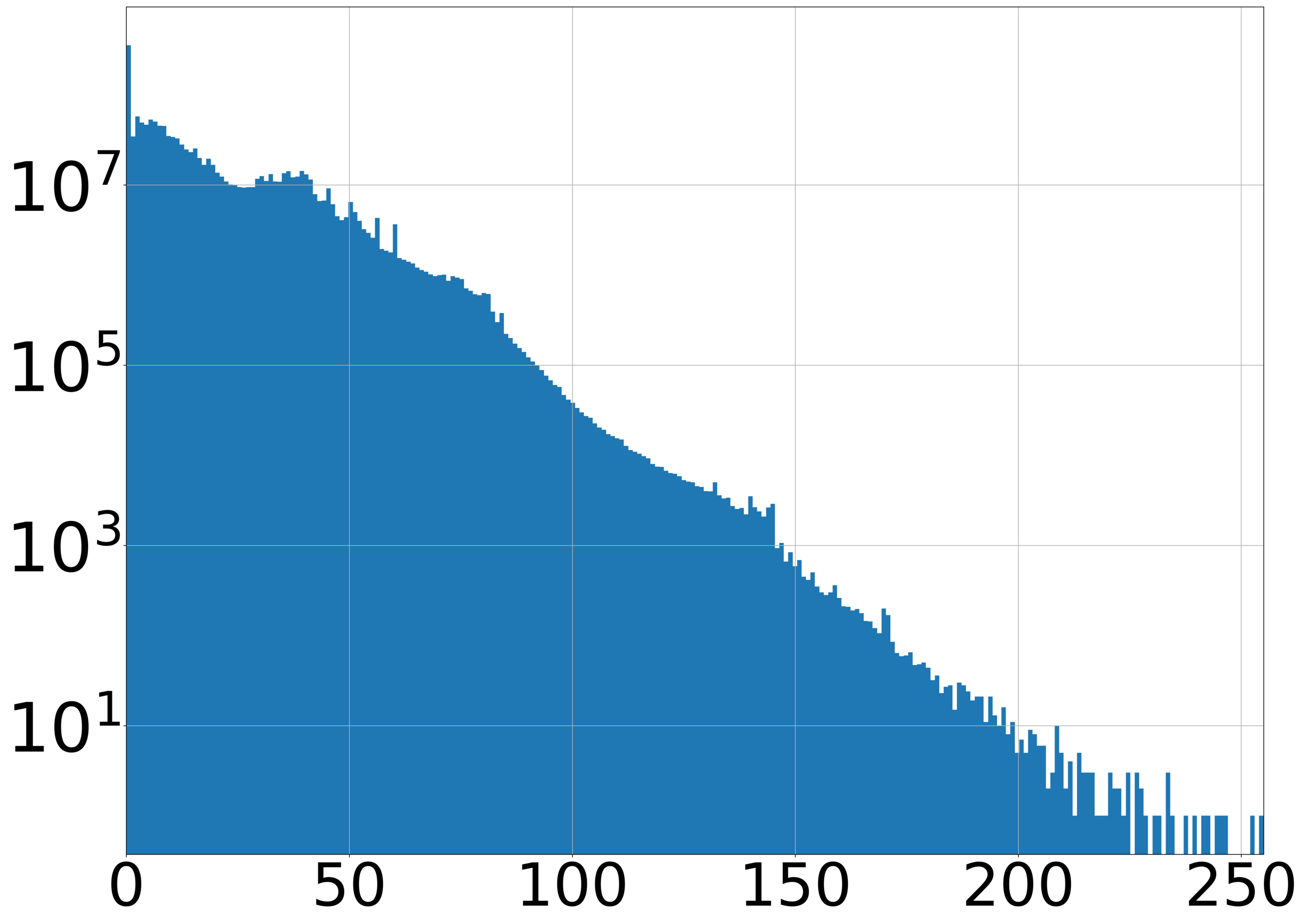}
  \end{subfigure}
  \begin{subfigure}{.19\textwidth}
    \centering
    \includegraphics[width=.9\linewidth]{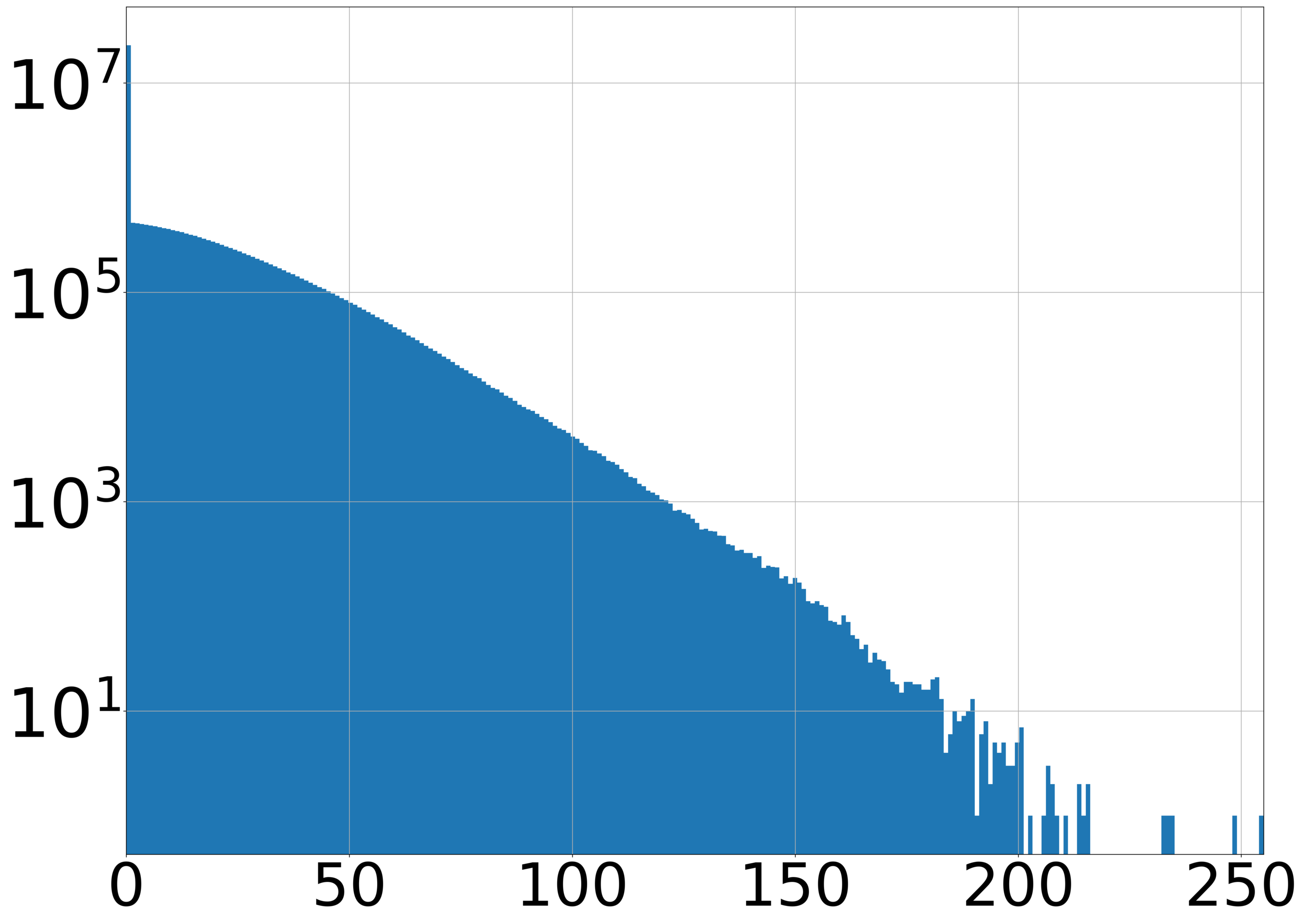}
  \end{subfigure}
  \begin{subfigure}{.19\textwidth}
    \centering
    \includegraphics[width=.9\linewidth]{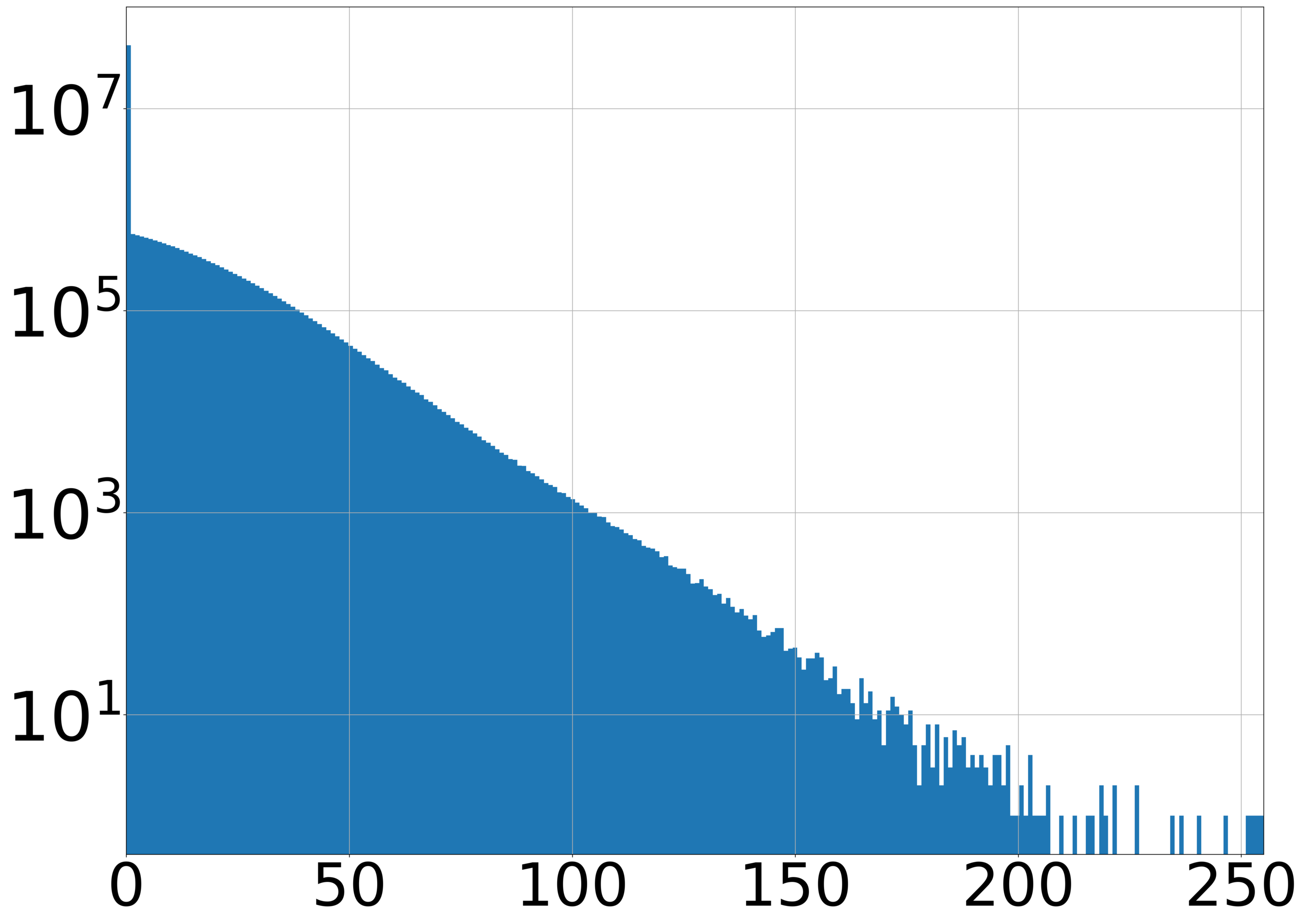}
  \end{subfigure}
  \begin{subfigure}{.19\textwidth}
    \centering
    \includegraphics[width=.9\linewidth]{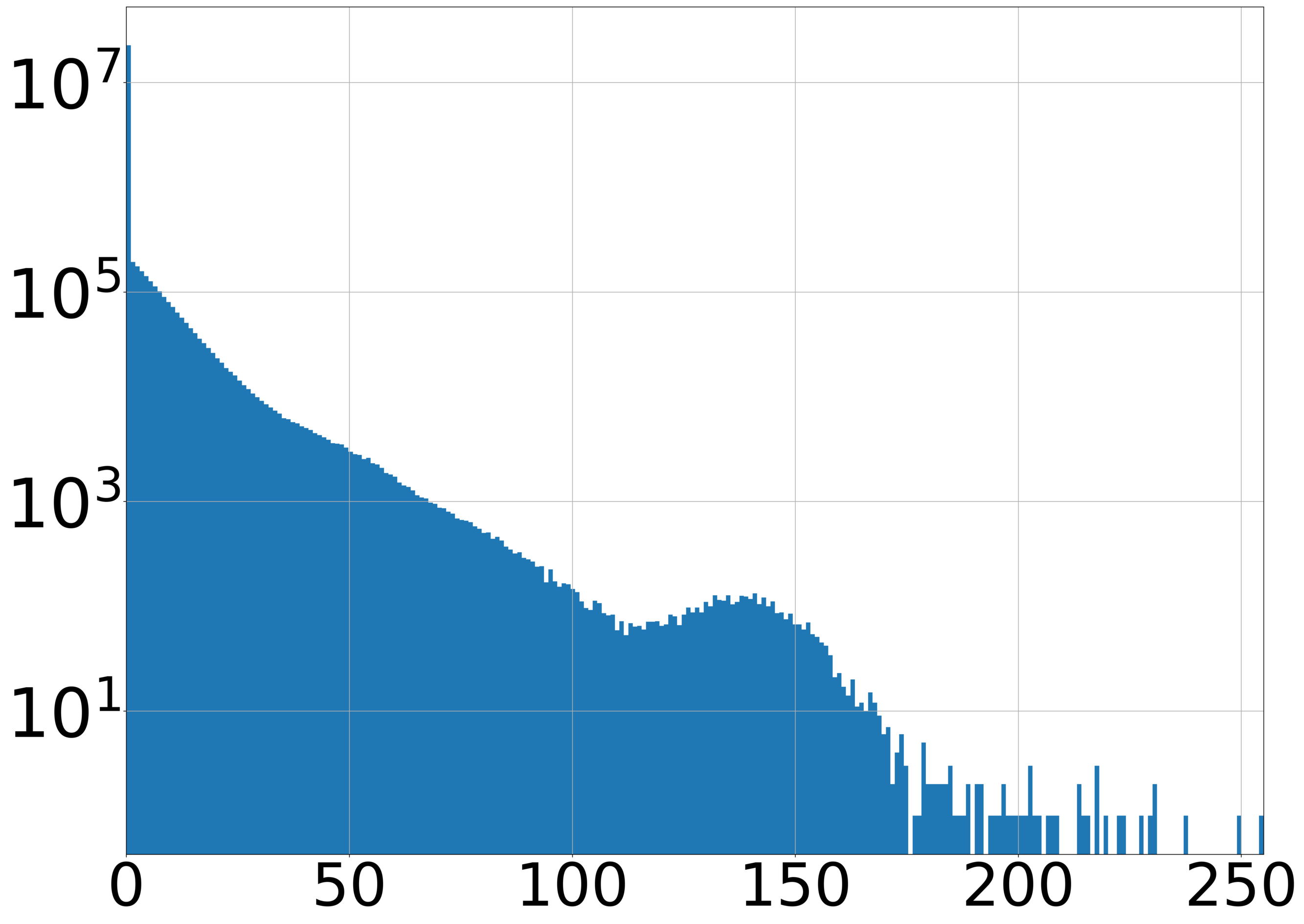}
  \end{subfigure}\\
  \begin{subfigure}{.19\textwidth}
    \centering
    \includegraphics[width=.9\linewidth]{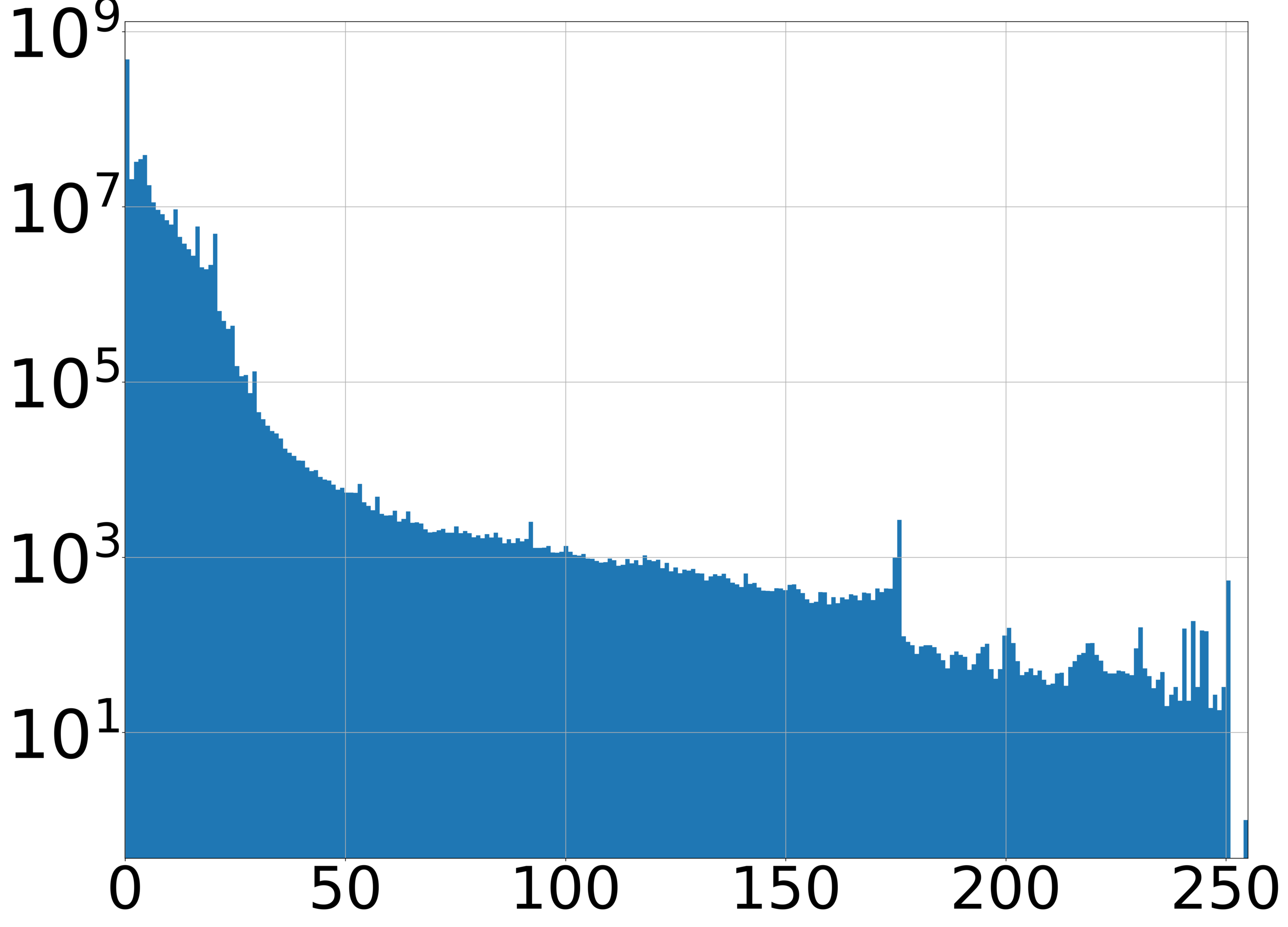}
    \caption{Conv2d\_1a\_3x3}
  \end{subfigure}%
  \begin{subfigure}{.19\textwidth}
    \centering
    \includegraphics[width=.9\linewidth]{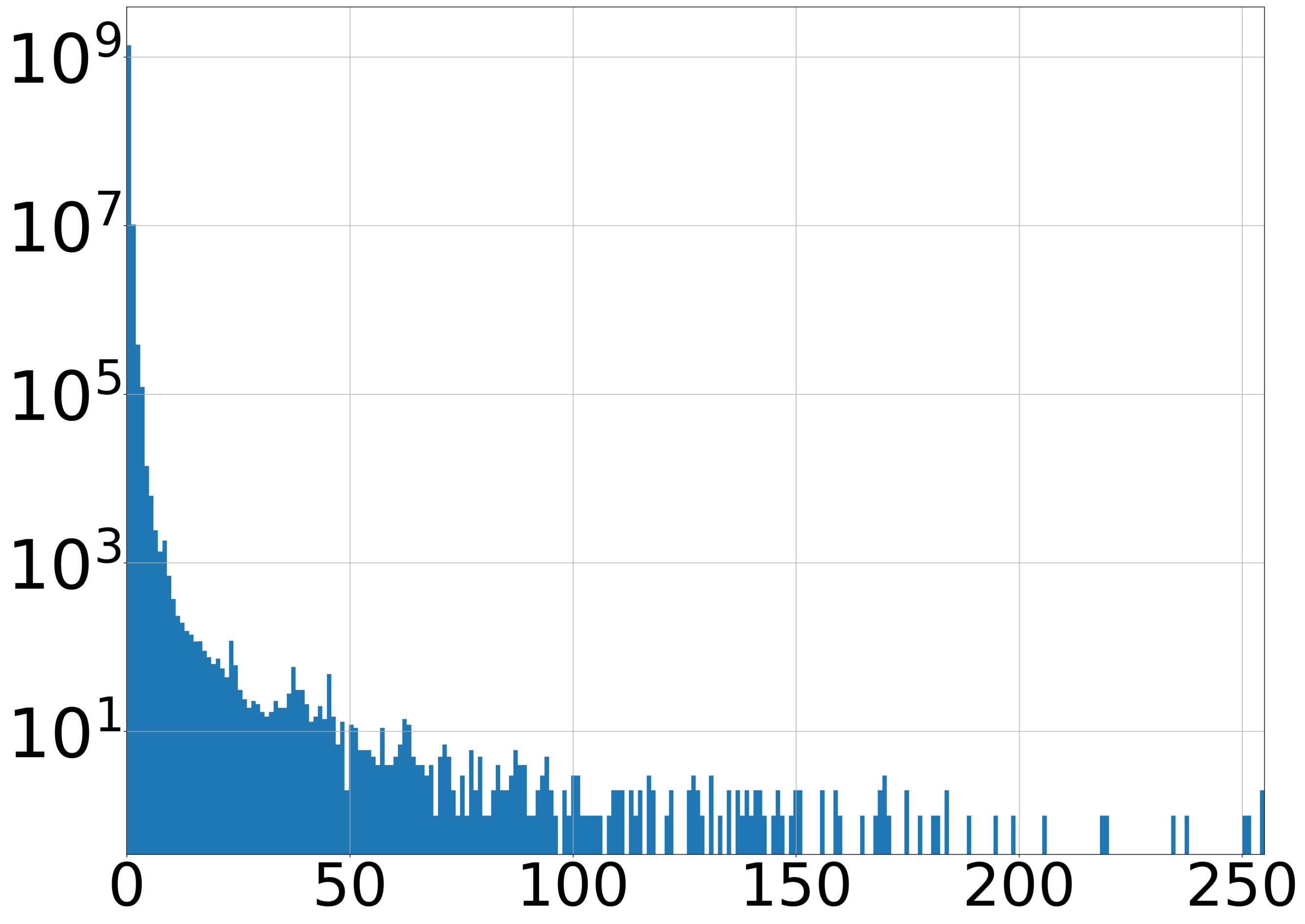}
    \caption{Conv2d\_2b\_3x3}
  \end{subfigure}
  \begin{subfigure}{.19\textwidth}
    \centering
    \includegraphics[width=.9\linewidth]{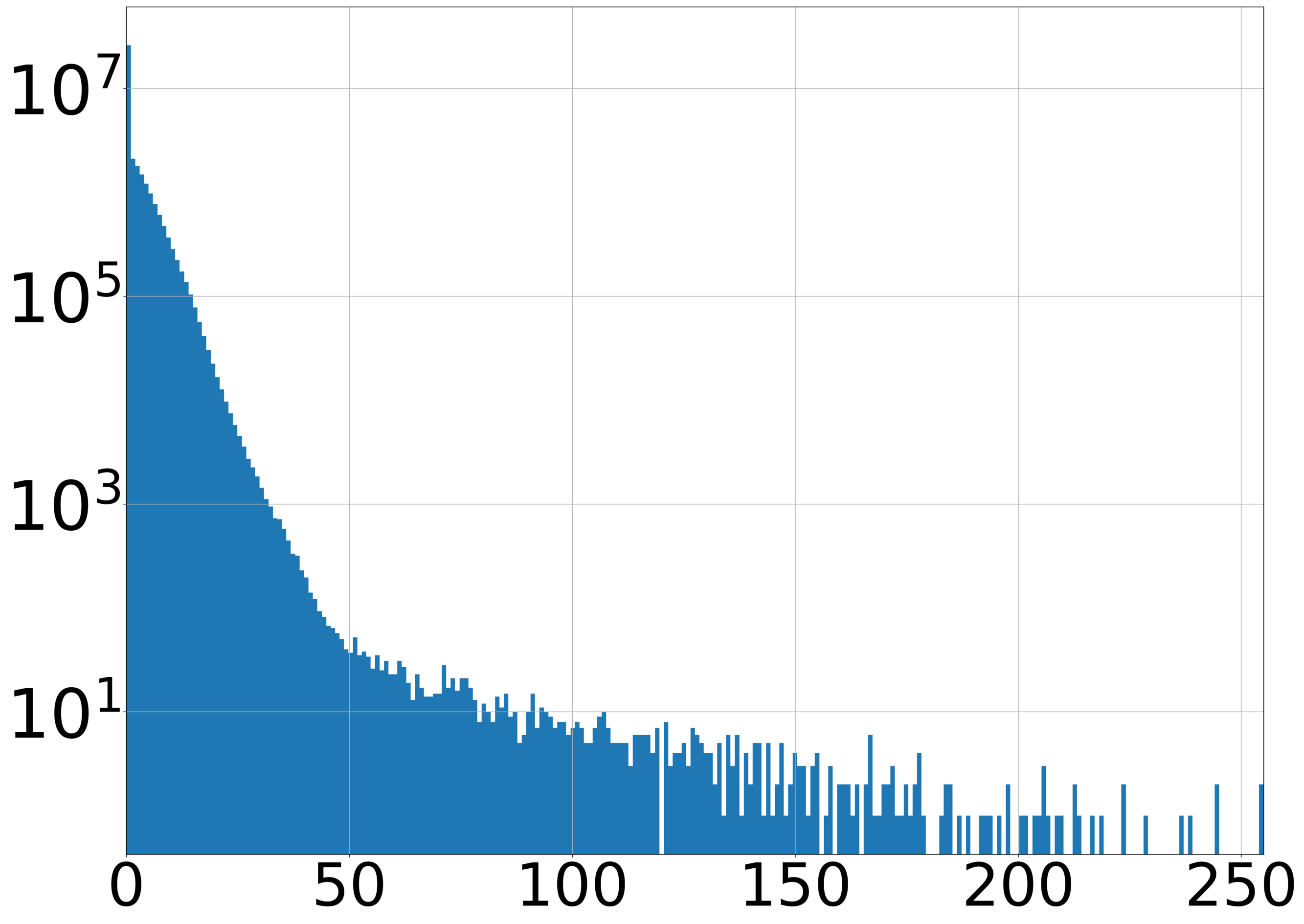}
    \caption{6b\_branch7x7dbl\_2}
  \end{subfigure}
  \begin{subfigure}{.19\textwidth}
    \centering
    \includegraphics[width=.9\linewidth]{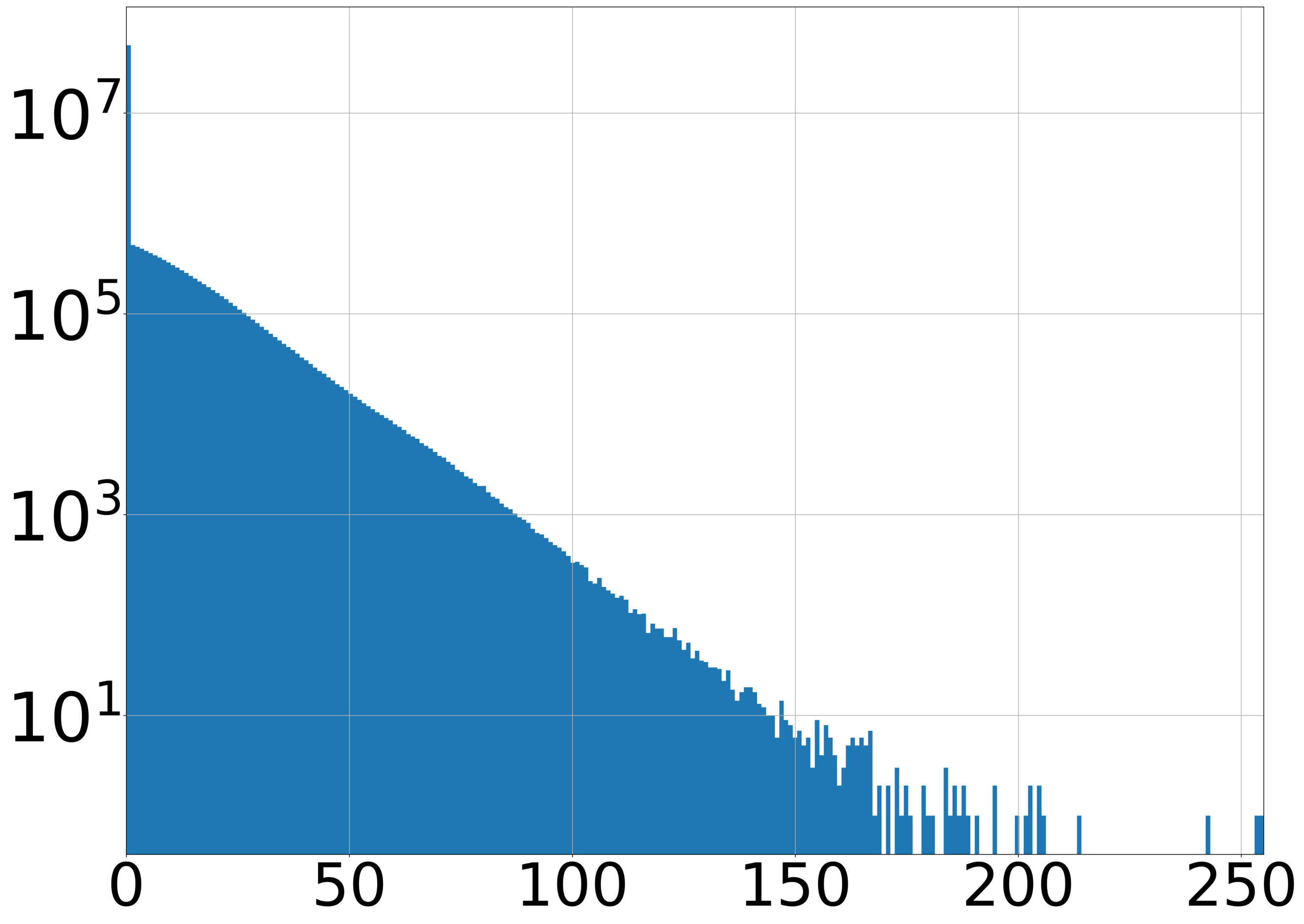}
    \caption{7a\_branch7x7x3\_3}
  \end{subfigure}
  \begin{subfigure}{.19\textwidth}
    \centering
    \includegraphics[width=.9\linewidth]{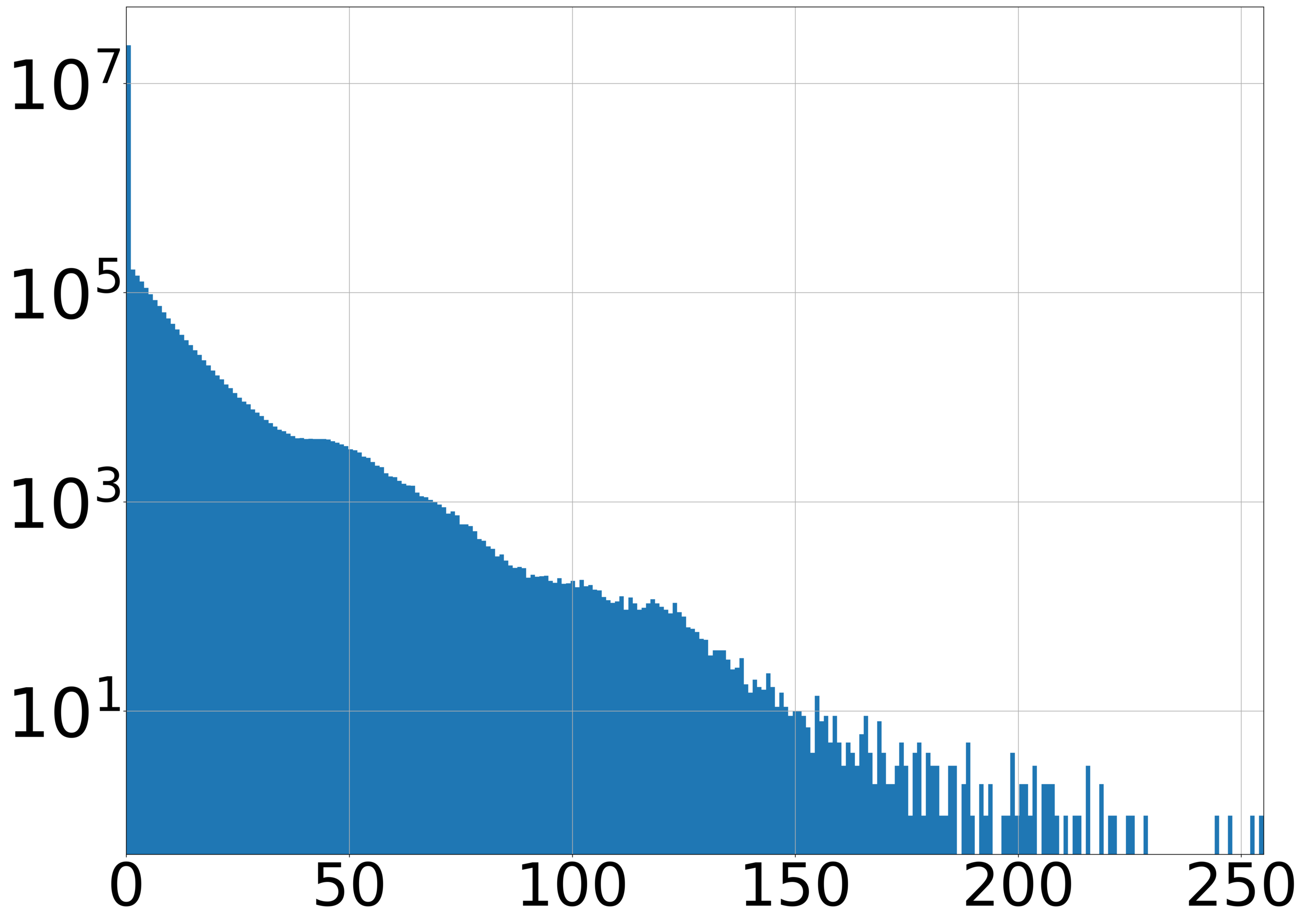}
    \caption{7b\_branch3x3dbl\_3a}
  \end{subfigure}\\
  \caption{Inception-V3 histograms of activation maps, before (above) and after (below) sparsification, in log-space from early (Conv2d\_1a\_3x3, Conv2d\_2b\_3x3), middle (6b\_branch7x7dbl\_2) and late (7a\_branch7x7x3\_3, 7b\_branch3x3dbl\_3a) layers, extracted from 1000 input images from the ImageNet (ILSVRC2012) \cite{Deng2009} training set. Activation maps are quantized to 8 bits (256 bins). The histograms show two important facts that sparse-exponential-Golomb takes advantage of: (1) Sparsity: Zero values are approximately one to two orders of magnitude more likely than any other value and (2) Long tail: Large values have a non-trivial probability. Sparsified activation maps not only have a higher percentage of zero values (enabling acceleration), but also have lower entropy (enabling compression).}
  \label{fig:hists-acts}
  \end{figure*}

The overwhelming majority of modern CNN architectures achieve sparsity in activation maps through the usage of ReLU as the activation function, which imposes a hard constraint on the intrinsic structure of the maps. We propose to aid training of neural networks by explicitly encoding in the cost function to be minimized, our desire to achieve sparser activation maps. We do so by placing a sparsity-inducing prior on $x_l$ for all layers, by modifying the cost function as follows:
\begin{equation}
  \label{eq:cost2}
  \begin{aligned}  
  E(w) & = && E_0 (w) + \frac{1}{N} \sum_{n=1}^{N} \sum_{l=0}^L \alpha_l \| x_{l,n} \|_1 \\
       & = && \frac{1}{N} \sum_{n=1}^{N} c'_n(w) + \lambda_w r ( w ),
  \end{aligned}
\end{equation}
where $\alpha_l \geq 0$ for $l=1,..\hdots,L-1$, $\alpha_0=\alpha_L=0$ and $c'_n$ is given by:

\begin{equation}
  \label{eq:cost3}
  c'_n(w) = c_n(w) + \sum_{l=0}^L \alpha_l \| x_{l,n} \|_1.
  \end{equation}
In Eq. \ref{eq:cost2} we use the $\boldsymbol{L}_1$ norm to induce sparsity on $x_l$ that acts as a proxy to the optimal, but difficult to optimize, $\boldsymbol{L}_0$ norm via a convex relaxation. The technique has been widely used in a variety of different applications including among others sparse coding \cite{olshausen1997} and LASSO \cite{tibshirani1996}.

While it is possible to train a neural network from scratch using the above cost function, our aim is to sparsify activation maps of existing state-of-the-art networks. Therefore, we modify the cost function from Eq. \ref{eq:cost1} to Eq. \ref{eq:cost2} during only the fine-tuning process of pre-trained networks. During fine-tuning, we backpropagate the gradients by computing the following quantities:

\begin{equation}
  \begin{aligned}
  \frac{\partial c'_n }{ \partial w_l } & =  \frac{\partial c'_n }{ \partial x_{l,n} } \cdot \frac{\partial x_{l,n} }{ \partial w_l } \\
  &=  \left[ \frac{ \partial c'_n }{ \partial x_{l+1,n} } \cdot \frac{\partial x_{l+1,n} }{ \partial x_{l,n} } + \frac{\partial c'_n }{ \partial x_{l,n} } \right] \cdot \frac{\partial x_{l,n} }{ \partial w_l }
  \end{aligned} ,
\end{equation}
where $w_l$ corresponds to the weights of layer $l$, $ {\partial c'_n }/{ \partial x_{l+1,n} }$ is the gradient backpropagated from layer $l+1$, $ {\partial x_{l+1,n} }/{ \partial x_{l,n} }$ is the output gradient of layer $l+1$ with respect to the input and the $j$-th element of ${\partial c_n' }/{ \partial x_{l,n}}$ is given by:
\begin{equation}
\begin{aligned}
  \frac{\partial c_n' }{ \partial x_{l,n}^j } = \alpha_l \frac{\partial \|x_{l,n}\|_1}{\partial x_{l,n}^j} = \left\{
                \begin{array}{ll}
                  +\alpha_l, &\text{~~if~~} x_{l,n}^j > 0\\
                  -\alpha_l, &\text{~~if~~} x_{l,n}^j < 0\\
                   0, &\text{~~if~~} x_{l,n}^j = 0
                \end{array},
              \right.
\end{aligned}
\end{equation}
where $x_{l,n}^j$ corresponds to the $j$-th element of (vectorized) $x_{l,n}$. Fig. \ref{fig:layerl} illustrates the computational graph and the flow of gradients for an example layer $l$. Note that $x_l$ can affect $c'$ both through $x_{l+1}$ and \emph{directly}, so we need to sum up both contributions during backpropagation.

\section{A Sparse Coding Interpretation}

Recently, connections between CNN's and Convolutional Sparse Coding (CSC) have been drawn \cite{papyan2017,sulam2017}. In a similar manner, it is also possible to interpret our proposed solution in Eq. \ref{eq:cost2} through the lens of sparse coding. Let us assume that for any given layer $l$, there exists an optimal underlying mapping $\hat{y_l}$ that we attempt to learn. Collectively $\{\hat{y}_l\}_{l=1,\hdots,L}$ define an optimal mapping from the input of the network to its output, while the network computes $\{ y_l = f_l(x_{l-1};w_l)\}_{l=1,\hdots,L}$. We can then think of training a neural network as an attempt to learn the optimal mappings $\hat{y}_l$ by minimizing the difference of the layer outputs:

\begin{equation}
  \label{eq:opt1}
  \begin{aligned}
  &\underset{w_l}{\text{minimize}} && \frac{1}{N} \sum_{n=1}^N \sum_{l=1}^{L}\|\hat{y}_{l,n} - y_{l,n}\|^2_2\\
  & \text{subject to}
  & & r(w_l) \leq C, l=1,\hdots,L.
  \end{aligned}
\end{equation}
In Eq. \ref{eq:opt1} we have also explicitly added the regularization term $r(w_l)$. $C \geq 0$ is a pre-defined constant that controls the amount of regularization. In the following, we restrict $r(w_l)$ to the $\boldsymbol{L}_2$ norm, $r(w_l) = \| w_l \|^2_2$, as it is by far the most common regularizer used and add the $\boldsymbol{L}_1$ prior on $x_l$:
\begin{equation} 
  \label{eq:opt2}
  \begin{aligned}    
    &\underset{w_l}{\text{minimize}} &&\frac{1}{N} \sum_{n=1}^N \left[ \sum_{l=1}^{L}\|\hat{y}_{l,n} - y_{l,n}\|^2_2 + \sum_{l=0}^{L} \alpha_l \| x_{l,n} \|_1 \right] \\
    & \text{subject to}
    & & \| w_l \|_2^2 \leq C, l=1,\hdots,L.
    \end{aligned}
\end{equation}
We can re-arrange the terms in Eq. \ref{eq:opt2} (and make use of the fact that $\alpha_L = 0$) as follows:
\begin{equation}
  \label{eqn:opt3}
  \begin{aligned}
  &\underset{w_l}{\text{minimize}} &&\frac{1}{N} \sum_{n=1}^N \sum_{l=1}^{L} P_{l,n} \\
  & \text{subject to}
  & & \| w_l \|_2^2 \leq C, l=1,\hdots,L
  \end{aligned}
\end{equation}
where:
\begin{equation}
P_{l,n} = \|\hat{y}_{l,n} - y_{l,n}\|^2_2 + \alpha_{l-1} \|x_{l-1,n}\|_1.
\end{equation}
For mappings $f_l(x_{l-1};w_l)$ that are linear, such as those computed by the convolutional and fully-connected layers, $f_l(x_{l-1};w_l)$ can also be written as a matrix-vector multiplication, by some matrix $W_l$, $y_l = W_lx_{l-1}$. $P_l$ can then be re-written as $P_l = \|\hat{y}_l - W_lx_{l-1}\|^2_2 + \alpha_{l-1} \|x_{l-1}\|_1$, which can be interpreted as a sparse coding of the optimal mapping $\hat{y}$. Therefore, Eq. \ref{eqn:opt3} amounts to computing the sparse coding representations of the pre-activation feature maps.

\section{Quantization}
\label{sec:quant}
We quantize floating point activation maps, $x_l$, to $q$ bits using linear (uniform) quantization:

\begin{equation}
  \label{eqn:quant_maps}
  x_l^{\text{quant}} = \frac{x_l - x_l^{\text{min}}}{x_l^{\text{max}} - x_l^{\text{min}} } \times ( 2^{q} - 1 ),
\end{equation}
where $x_l^{\text{min}}=0$ and $x_l^{\text{max}}$ corresponds to the maximum value of $x_l$ in layer $l$ in the training set. Values above $x_l^{\text{max}}$ in the testing set are clipped. While we do not retrain our models after quantization, it has been shown by the literature to improve model accuracy \cite{cai2017,lin2016fixed}. We also believe that a joint optimization scheme where we simultaneously perform quantization and sparsification can further improve our results and leave it for future work.

\section{Entropy Coding}
\label{sec:enc}

A number of different schemes have been devised to store sparse matrices effectively. Compressed sparse row (CSR) and compressed sparse column (CCS) are two representations that strike a balance between compression and efficiency of arithmetic operation execution. However, such methods assume that the entire matrix is available prior to storage, which is not necessarily true in all of our use cases. For example, in neural network hardware accelerators data are often streamed as they are computed and there is a great need to compress in an on-line fashion. Hence, we shift our focus to algorithms that can encode one element at a time. 
We present a new entropy coding algorithm, sparse-exponential-Golomb (SEG) (Alg. \ref{alg:sparse_exp_golomb}) that is based on the effective exponential-Golomb (EG) \cite{teuhola1978,wen1998}. SEG leverages on two facts: (1) Most activation maps are sparse and, (2) The first-order probability distribution of the activation maps has a long tail.

Exponential-Golomb is most commonly used in H.264 \cite{h264} and HEVC \cite{hevc} video compression standards with $k=0$ as a standard parameter. $k=0$ is a particularly effective parameter value when data is sparse since it assigns a code word of length 1 for the value $x=0$. However, it also requires that the probability distribution of values falls-off rapidly. Activation maps have a non-trivial number of large values (especially for $q=8,12,16$), rendering $k=0$ ineffective. Fig. \ref{fig:hists-acts} demonstrates this fact by showing histograms of activation maps for a sample of layers from Inception-V3. While this issue could be solved by using larger values of $k$, the consequence is that the value $x=0$ is no longer encoded with a 1 bit code word (see Table \ref{tab:code_length}). SEG solves this problem by dedicating the code word `1' for $x=0$ and by pre-appending the code word generated by EG with a `0' for $x>0$. Sparse-exponential golomb can be found in Alg. \ref{alg:sparse_exp_golomb} while exponential-Golomb \cite{teuhola1978} is provided in Appendix \ref{sec:exp-golomb}.

\begin{algorithm}[t]  
  \caption{Sparse-exponential-Golomb}
  \begin{algorithmic}  
  \STATE \textbf{Input}: Non-negative integer $x$, Order $k$\\
  \STATE \textbf{Output}: Bitstream $y$\\  
  $\text{function \textbf{encode\_sparse\_exp\_Golomb} }( x , k )$\\
  \{\\
  \INDSTATE If $k == 0$:\\
  \INDSTATE[2] $y=\text{encode\_exp\_Golomb} ( x , k )$
  \INDSTATE Else:\\
  \INDSTATE[2] If $x == 0$:\\
  \INDSTATE[4] Let $y=\text{`1'}$\\
  \INDSTATE[2] Else:\\
  \INDSTATE[4] Let $y = \text{`0'} + \text{encode\_exp\_Golomb} ( x - 1 , k )$ \\
  \INDSTATE Return $y$\\   
  \}\\
  \STATE \textbf{Input}: Bitstream $x$, Order $k$\\  
  \STATE \textbf{Output}: Non-negative integer $y$ \\
  $\text{function \textbf{decode\_sparse\_exp\_Golomb} }( x , k )$\\
  \{\\
  \INDSTATE If $k == 0$:\\
  \INDSTATE[2] $y = \text{decode\_exp\_Golomb} ( x , k )$
  \INDSTATE Else:\\
  \INDSTATE[2] If $x[0] == \text{`1'}$:\\
  \INDSTATE[4] Let $y=0$\\
  \INDSTATE[2] Else:\\
  \INDSTATE[4] Let $y = 1 + \text{decode\_exp\_Golomb} ( x[1\colon] , k )$ \\
  \INDSTATE Return $y$\\   
  \}\\

 \end{algorithmic}
 \label{alg:sparse_exp_golomb}
 \end{algorithm}
 \begin{table}[t]
  \footnotesize
  \centering
  \begin{tabular}{ccccc}
  \toprule
  \textbf{Algorithm} & $k=0$ & $k=4$ & $k=8$ & $k=12$\\
  \midrule
  EG & 1 & 5 & 9 & 13\\
  \midrule
  SEG & 1 & 1 & 1 & 1\\
  \bottomrule    
\end{tabular}
\caption{Code word length comparison of $x=0$ between EG and SEG for different values of $k$.}
\label{tab:code_length} 
\end{table}

\section{Experiments}

\begin{table*}[t]
  \centering
  \resizebox{0.9\linewidth}{!}
  {
  \begin{tabular}{c|cccccc}
  \toprule
  \textbf{Dataset} &\textbf{Model} & \textbf{Variant} &\textbf{Top-1 Acc.}  & \textbf{Top-5 Acc.} & \textbf{Acts. (\%)} & \textbf{Speed-up}\\
  \midrule
  \multirow{ 2}{*}{MNIST}  & \multirow{ 2}{*}{LeNet-5}& Baseline & 98.45\% & - & 53.73\% & 1.0$\times$\\
   & & Sparse & \makecell{\textbf{98.48\% (+0.03\%)}} & - & \textbf{23.16\%} & \textbf{2.32$\times$}\\
  \midrule
  \multirow{ 2}{*}{CIFAR-10}  & \multirow{ 2}{*}{MobileNet-V1} & Baseline & 89.17\% & - & 47.44\% & 1.0$\times$\\
  & & Sparse & \textbf{89.71\% (+0.54\%)} & - & \textbf{29.54\%} & \textbf{1.61$\times$}\\  
  \midrule
  \multirow{ 10}{*}{ImageNet}  & \multirow{ 3}{*}{Inception-V3} & Baseline & 75.76\% & 92.74\% & 53.78\% & 1.0$\times$\\
  & & Sparse & \textbf{76.14\% (+0.38\%)}  & \textbf{92.83\% (+0.09\%)} & 33.66\% & 1.60$\times$\\
  & & Sparse\_v2 & 68.94\% (-6.82\%) & 88.52\% (-4.22\%) & \textbf{25.34\%} & \textbf{2.12$\times$}\\       
  \cmidrule(lr{1em}){2-7}
    & \multirow{ 3}{*}{ResNet-18} & Baseline & 69.64\% & 88.99\% & 60.64\% & 1.0$\times$\\
  & & Sparse & \textbf{69.85\% (+0.21\%)}  & \textbf{89.27\% (+0.28\%)} & 49.51\% & 1.22$\times$\\
  & & Sparse\_v2 & 68.62\% (-1.02\%) & 88.41\% (-0.58\%) & \textbf{34.29\%} & \textbf{1.77$\times$}\\
  \cmidrule(lr{1em}){2-7}
    & \multirow{ 3}{*}{ResNet-34} & Baseline & 73.26\% & 91.43\% & 57.44\% & 1.0$\times$\\
  & & Sparse & \textbf{73.95\%(+0.69\%)}  & \textbf{91.61\% (+0.18\%)} & 46.85\% & 1.23$\times$\\
  & & Sparse\_v2 & 67.73\% (-5.53\%) & 87.93\% (-3.50\%) & \textbf{29.62\%} & \textbf{1.94$\times$}\\
  \bottomrule    
\end{tabular}
  }
\caption{Accelerating neural networks via sparsification. Numbers in brackets indicate change in accuracy. Acts. (\%) shows the percentage of non-zero activations.}
\label{tab:spars-results} 
\end{table*} 

\begin{table}[t]
  \centering
  \resizebox{\columnwidth}{!}
  {  
  \begin{tabular}{c|cccc}    
  \toprule
  \textbf{Network} & \textbf{Algorithm} & \textbf{Top-1 Acc. Change} & \textbf{Top-5 Acc. Change} & \textbf{Speed-up}\\
  \midrule
  \multirow{5}{*}{ResNet-18} & Ours (Sparse)& +0.21\% & +0.28\% & 18.4\%\\
  & \textbf{Ours (Sparse\_v2)} & \textbf{-1.02\%} & \textbf{-0.58\%} & \textbf{43.5\%}\\
   & LCCL \cite{dong2017} & -3.65\% & -2.30\% & 34.6\%\\
   & BWN \cite{rastegari2016} & -8.50\% & -6.20\% & 50.0\%\\
   & XNOR \cite{rastegari2016} & -18.10\% & -16.00\% & 98.3\%\\
  \midrule
  \multirow{4}{*}{ResNet-34} & Ours (Sparse) & {+0.69\%} & {{+0.18\%}} & 18.4\%\\
   & \textbf{Ours (Sparse\_v2)}  & \textbf{-5.53\%} & \textbf{-3.50\%} & \textbf{48.4}\%\\
   & LCCL \cite{dong2017} & {-0.43\%}  & {-0.17\%} & 24.8\%\\     
   & PFEC \cite{Li2016} & {-1.06\%} & - & 24.2\%\\    
   \midrule
   \multirow{3}{*}{LeNet-5} & \textbf{Ours (Sparse)} & \textbf{+0.03\%} & - & \textbf{56.9\%}\\
   & \cite{han2015} ($p=70\%$) & -0.12\%  & - & 7.3\%\\
   & \cite{han2015} ($p=80\%$) & -0.57\%  & - & 14.7\%\\
   \bottomrule       
\end{tabular}
}
\caption{Comparison between various state-of-the-art acceleration methods on ResNet-18/34 and LeNet-5. For ResNet-18/34, we emphasized in bold Sparse\_v2 as a good compromise between acceleration and accuracy. $p=70\%$ indicates that the pruned network has $30\%$ non-zero weights. Speed-up calculation follows the convention from \cite{dong2017}, where it is reported as 1 - (non-zero activations of sparse model) / (non-zero activations of baseline).}
\label{tab:comp-spars} 
\end{table}

In the experimental section, we investigate two important applications: (1) acceleration of computation and (2) compression of activation maps. We carry our experiments on three different datasets, MNIST \cite{lecun98}, CIFAR-10 \cite{krizhevsky2009learning} and ImageNet ILSVRC2012 \cite{Deng2009} and five different networks, a LeNet-5 \cite{lecun98} variant\footnote{\url{https://github.com/pytorch/examples/}}, MobileNet-V1 \cite{howard2017}, Inception-V3 \cite{szegedy2016}, ResNet-18 \cite{he2016} and ResNet-34 \cite{he2016}. These networks cover a wide variety of network sizes, complexity in design and efficiency in computation. For example, unlike AlexNet \cite{krizhevsky2012} and VGG-16 \cite{simonyan2014} that are over-parameterized and easy to compress, MobileNet-V1 is an architecture that is already designed to reduce computation and memory consumption and as a result, it presents a challenge for achieving further savings. Furthermore, compressing and accelerating Inception-V3 and the ResNet architectures has tremendous usefulness in practical applications since they are among the state-of-the-art image classification networks. 

\begin{figure}[t]
  \begin{subfigure}{0.5\columnwidth}
    \centering
    \includegraphics[width=.95\linewidth]{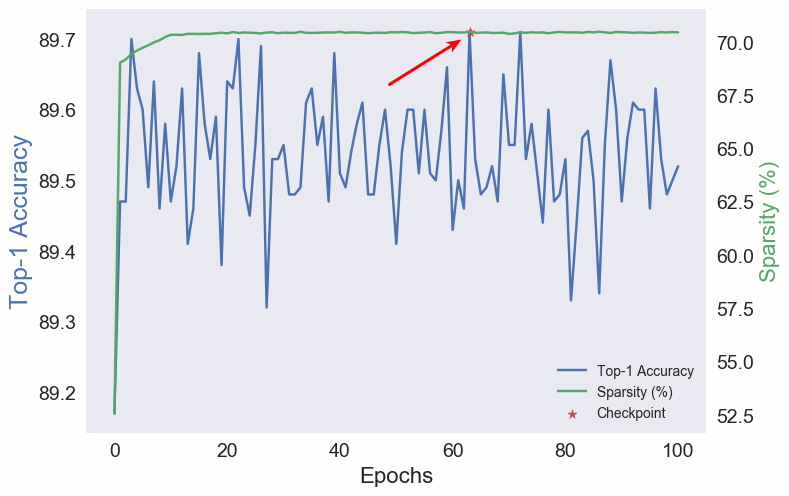}
    \caption{MobileNet-V1}
  \end{subfigure}%
  \begin{subfigure}{0.5\columnwidth}
    \centering
    \includegraphics[width=.95\linewidth]{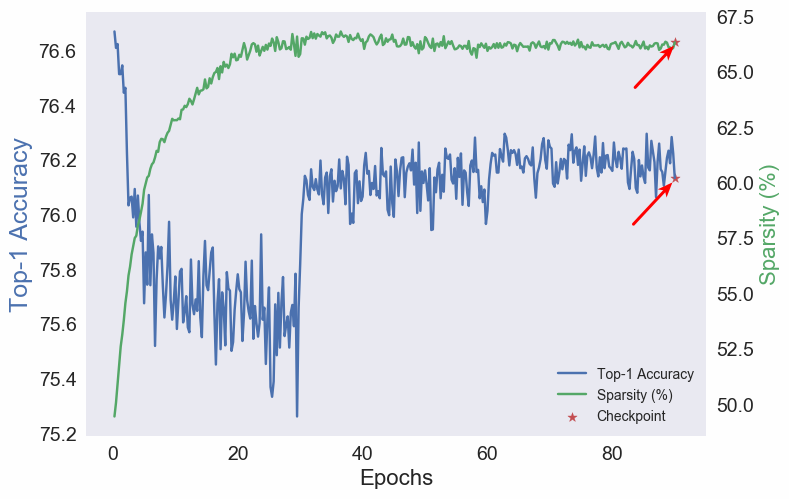}
    \caption{Inception-V3}
  \end{subfigure}  
  \caption{Evolution of Top-1 accuracy and activation map sparsity of MobileNet-V1 and Inception-V3 during training on the validation sets of CIFAR-10 and ILSVRC2012 respectively. The red arrow indicates the checkpoint selected for the sparse model.}
  \label{fig:train-curves}
  \end{figure}

\textbf{Acceleration.} In Table \ref{tab:spars-results}, we summarize our speed-up results. Baselines were obtained as follows: LeNet-5 and MobileNet-V1\footnote{\url{https://github.com/kuangliu/pytorch-cifar/}} were trained from scratch, while Inception-V3 and ResNet-18/34 were obtained from the PyTorch \cite{paszke2017automatic} repository\footnote{\url{https://pytorch.org/docs/stable/torchvision/}}. The results reported were computed on the validation sets of the corresponding datasets. The speed-up factor is calculated by dividing the number of non-zero activations of the baseline by the number of non-zero activations of the sparse models. For Inception-V3 and ResNet-18/34, we present two sparse variants, one targeting minimum reduction in accuracy (sparse) and the other targeting high sparsity (sparse\_v2).

\begin{figure*}[t]
  \begin{subfigure}{0.19\linewidth}
    \centering
    \includegraphics[width=0.99\linewidth]{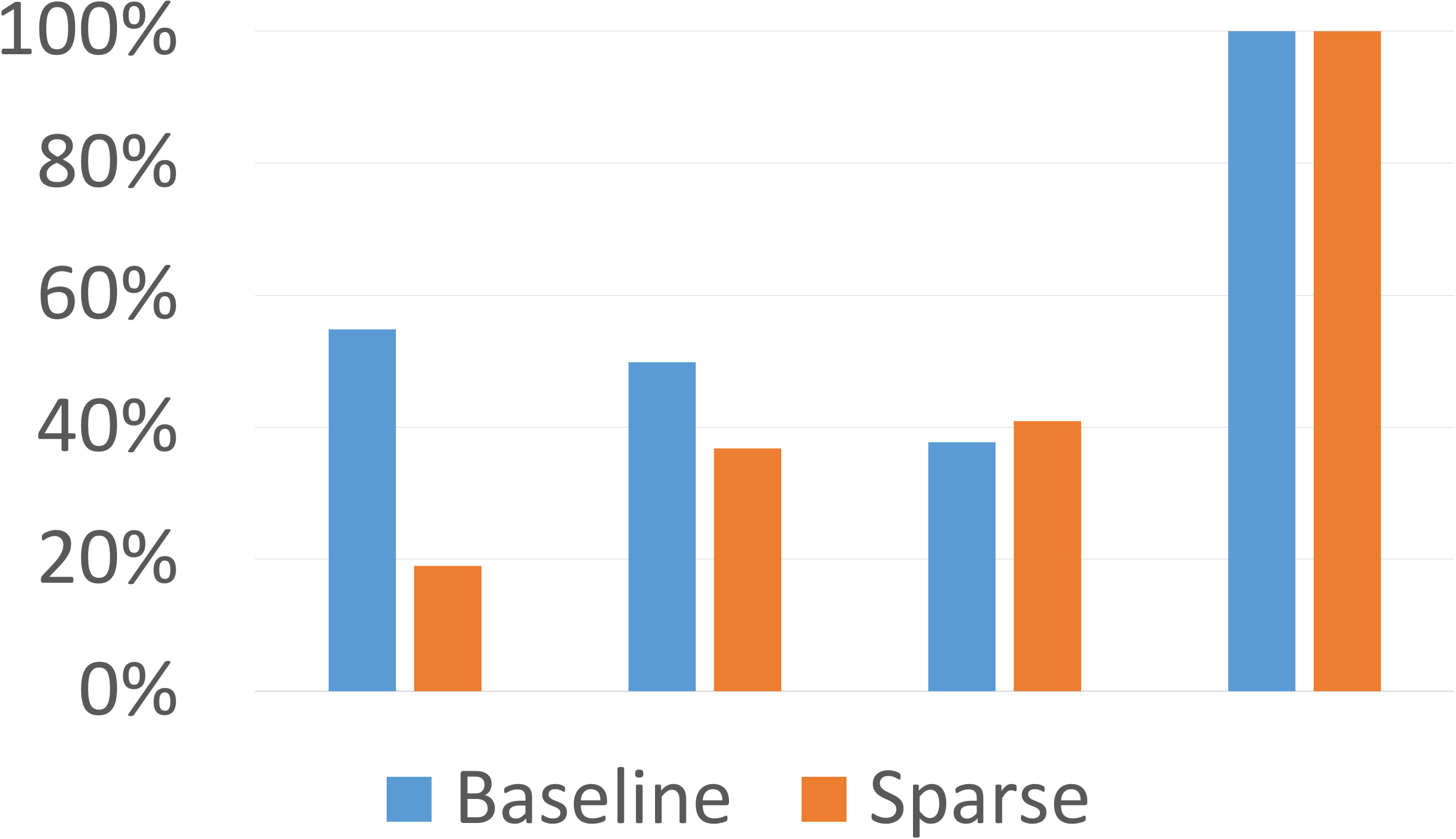}
  \end{subfigure}%
  \begin{subfigure}{0.19\linewidth}
    \centering
    \includegraphics[width=0.99\linewidth]{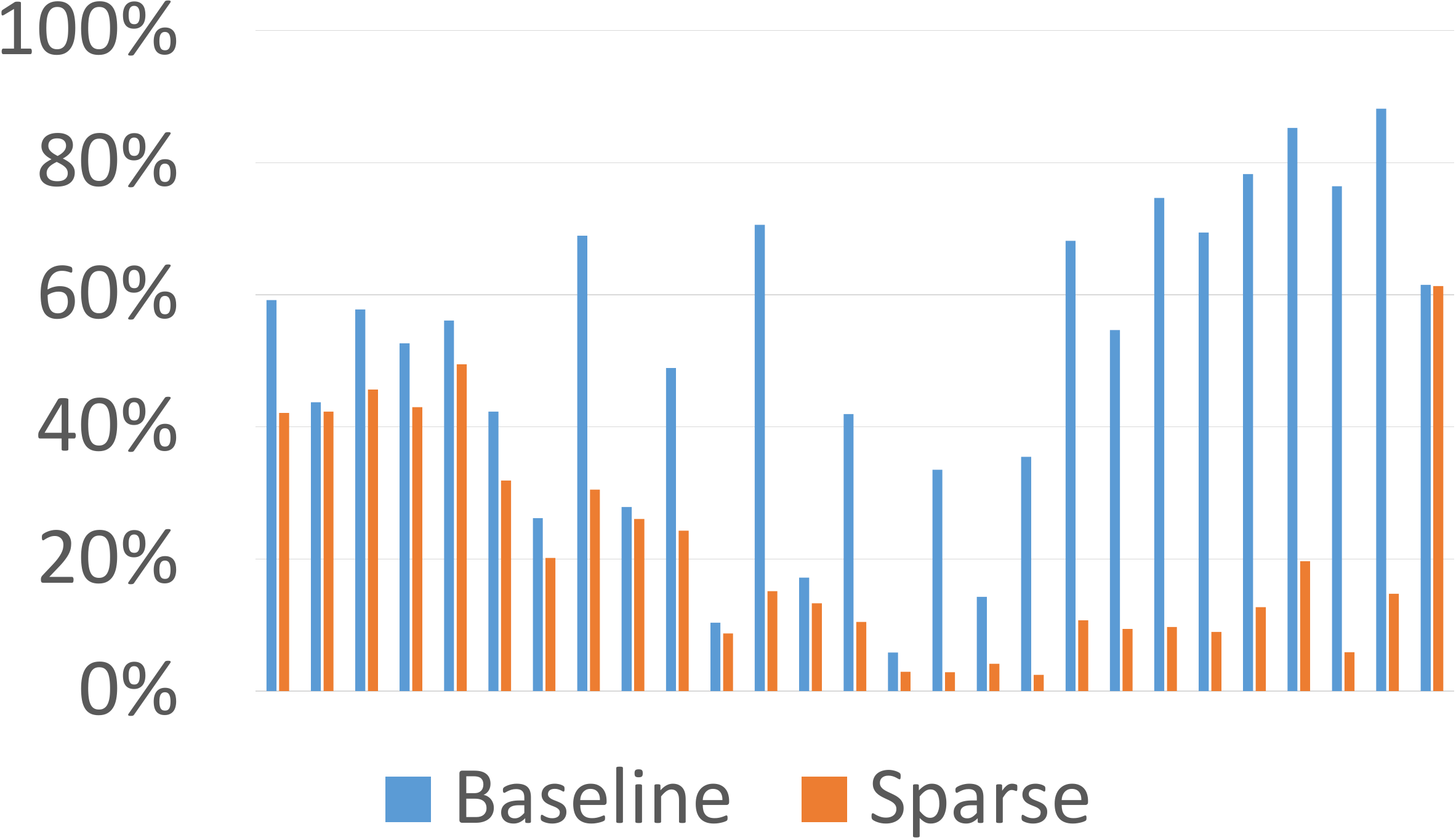}
  \end{subfigure}%
  \begin{subfigure}{0.19\linewidth}
    \centering
    \includegraphics[width=0.99\linewidth]{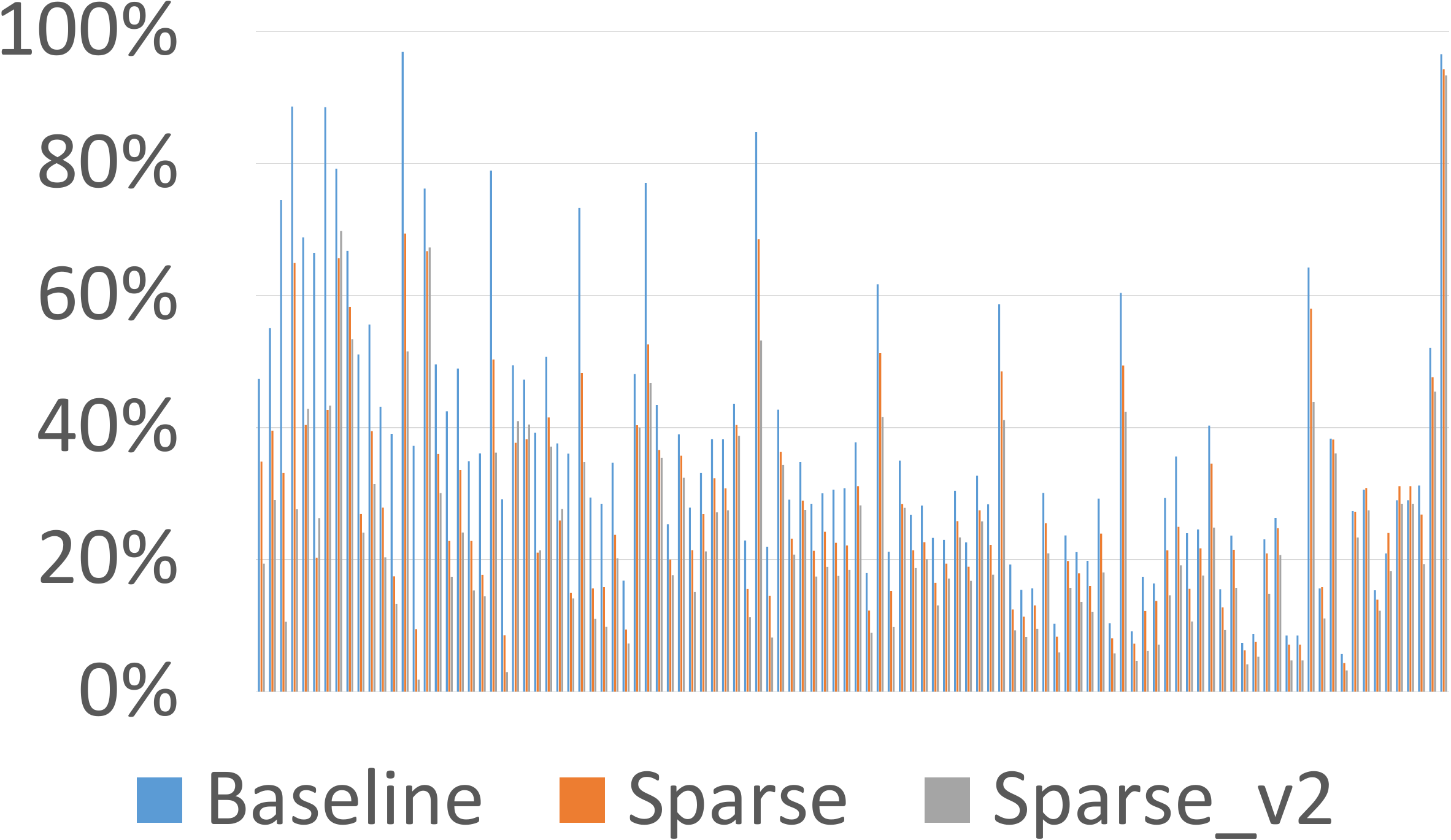}
  \end{subfigure}%
  \begin{subfigure}{0.19\linewidth}
    \centering
    \includegraphics[width=0.99\linewidth]{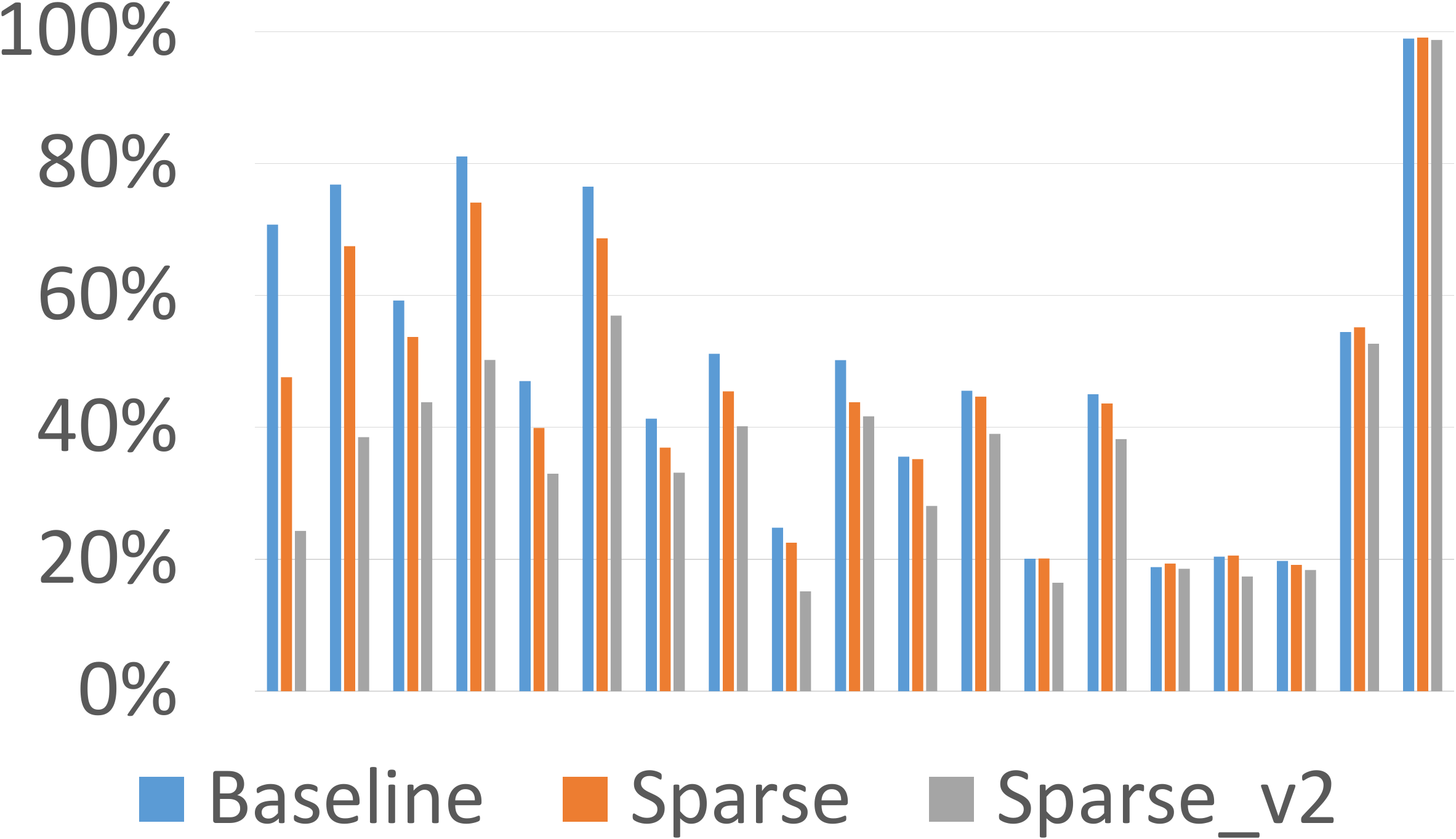}
  \end{subfigure}%
  \begin{subfigure}{0.19\linewidth}
    \centering
    \includegraphics[width=0.99\linewidth]{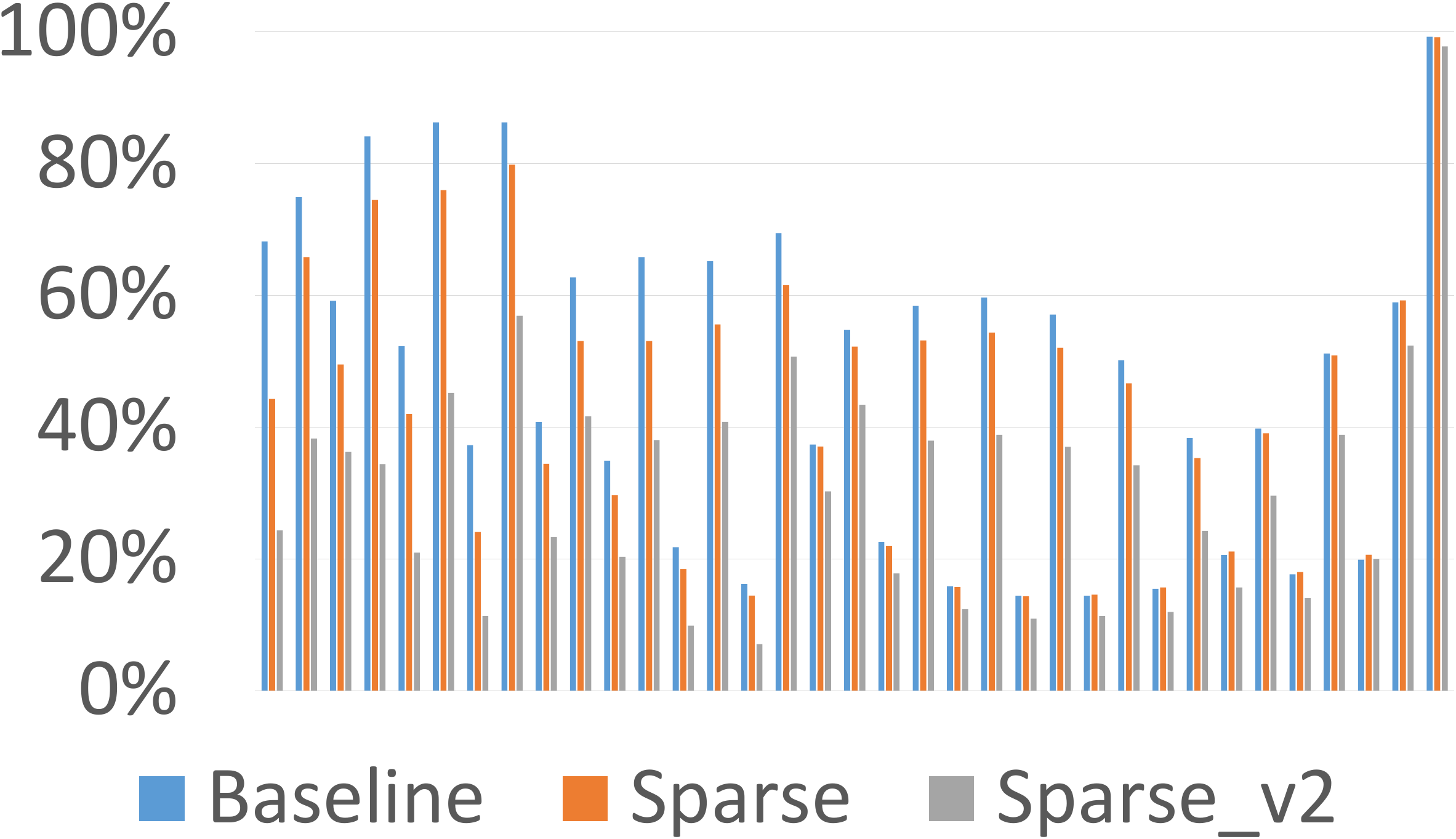}
  \end{subfigure}\\
  \begin{subfigure}{0.19\linewidth}
    \centering
    \includegraphics[width=0.99\linewidth]{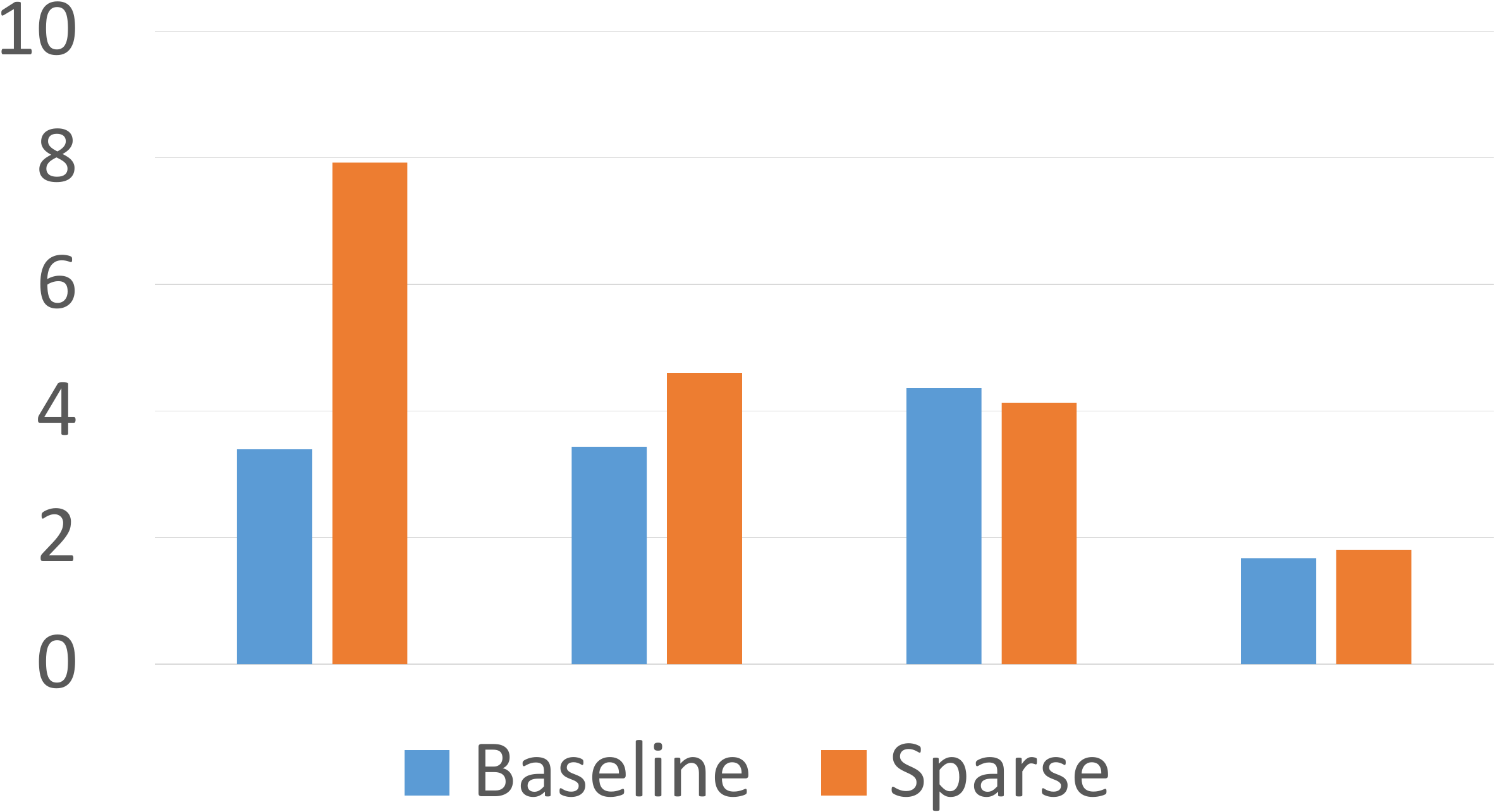}
    \caption{LeNet-5}
  \end{subfigure}%
  \begin{subfigure}{0.19\linewidth}
    \centering
    \includegraphics[width=0.99\linewidth]{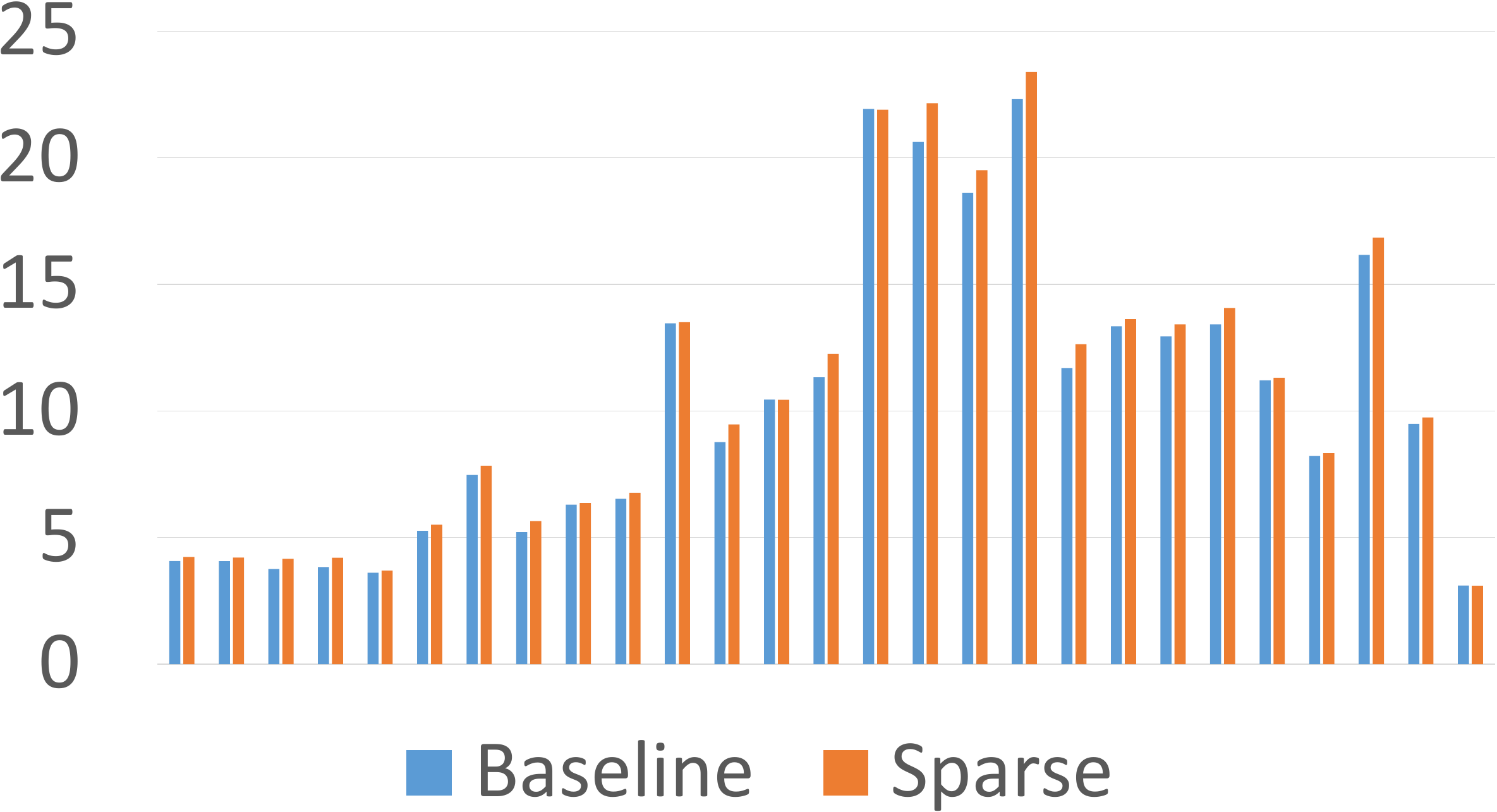}
    \caption{MobileNet-V1}
  \end{subfigure}%
  \begin{subfigure}{0.19\linewidth}
    \centering
    \includegraphics[width=0.99\linewidth]{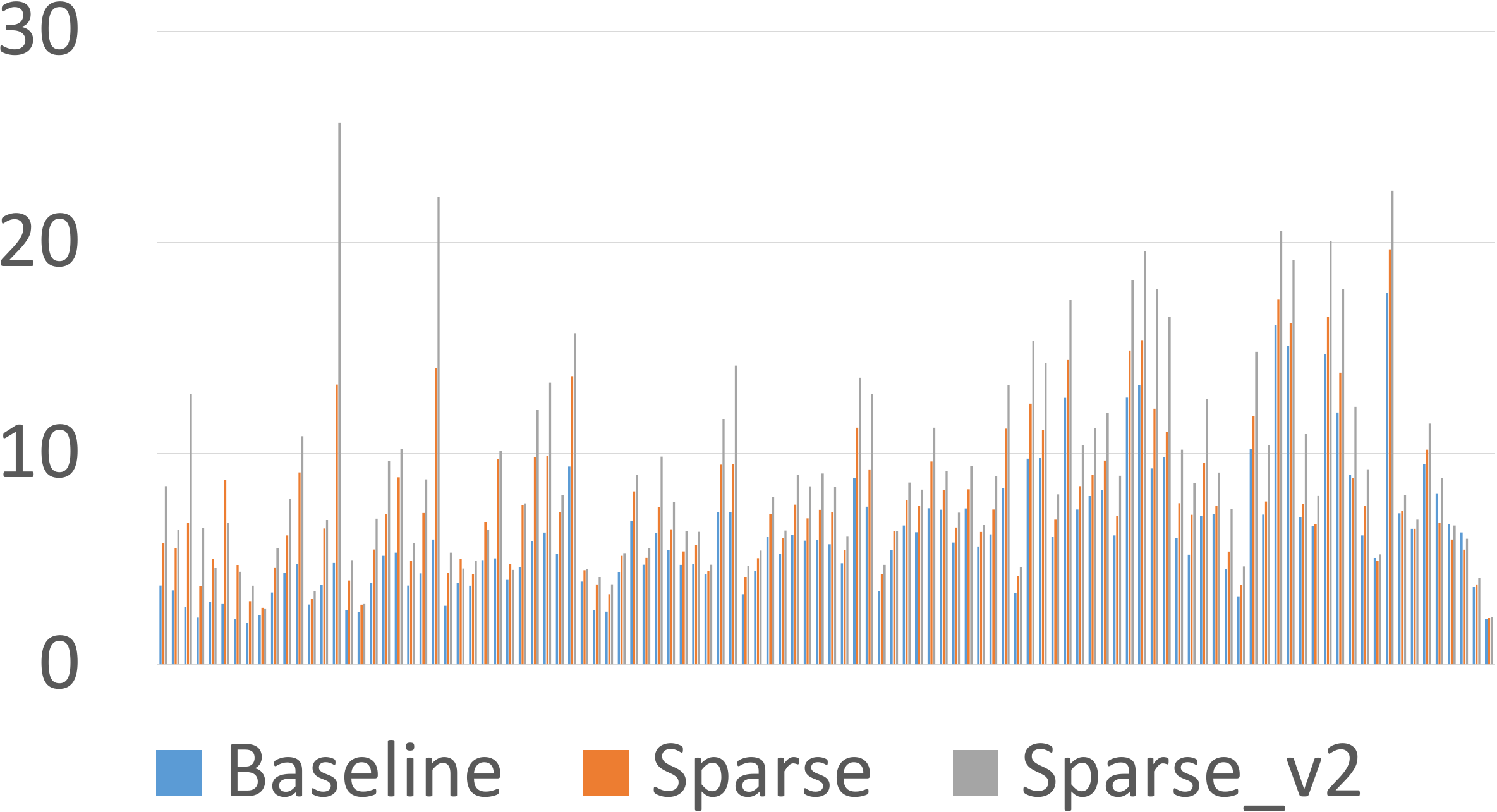}
    \caption{Inception-V3}
  \end{subfigure}%
  \begin{subfigure}{0.19\linewidth}
    \centering
    \includegraphics[width=0.99\linewidth]{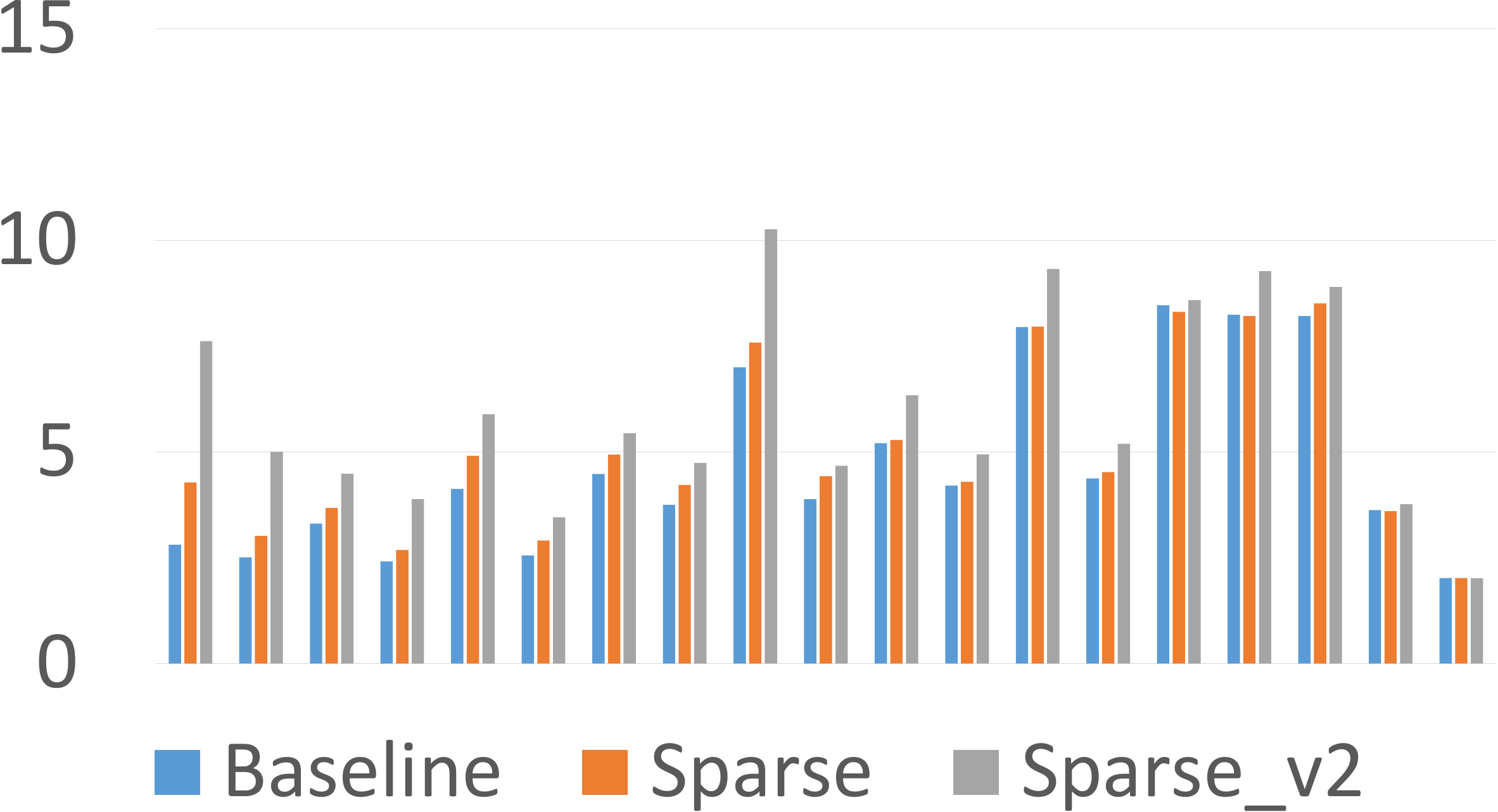}
    \caption{ResNet-18}
  \end{subfigure}%
  \begin{subfigure}{0.19\linewidth}
    \centering
    \includegraphics[width=0.99\linewidth]{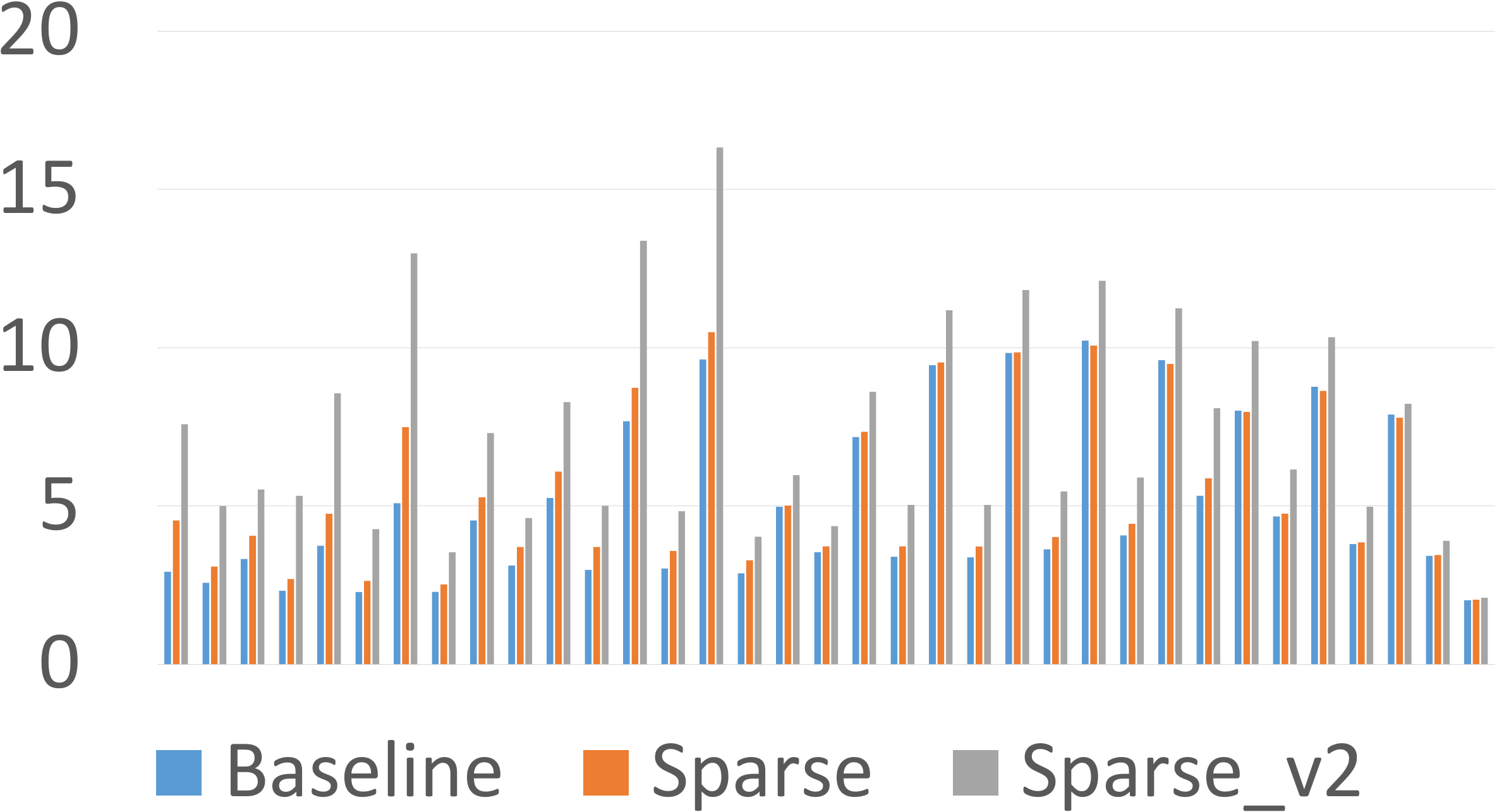}
    \caption{ResNet-34}
  \end{subfigure}%        
  \caption{Percentage of non-zero activations (above) and compression gain (below) per layer for various network architectures before and after sparsification.}
  \label{fig:layer-sparsity}
\end{figure*}

Many of our sparse models not only have an increased sparsity in their activation maps, but also demonstrate increased accuracy. This alludes to the well-known fact that sparse activation maps have strong representational power \cite{glorot2011} and the addition of the sparsity-inducing prior does not necessarily trade-off accuracy for sparsity. When the regularization parameter is carefully chosen, the data term and the prior can work together to improve both the Top-1 accuracy and the sparsity of the activation maps. \emph{Inception-V3 can be accelerated by as much as 1.6$\times$ with an increase in accuracy of $0.38\%$, while ResNet-18 achieves a speed-up of 1.8$\times$ with a modest $1\%$ accuracy drop. LeNet-5 can be accelerated by 2.3$\times$, MobileNet-V1 by 1.6$\times$ and finally ResNet-34 by 1.2$\times$, with all networks exhibiting accuracy increase.} The regularization parameters, $\alpha_l$, can be adapted for each layer independently or set to a common constant. We experimented with various configurations and the selected parameters are shared in Appendix \ref{sec:regparam}. To determine the selected $\alpha_l$, we used grid-based hyper-parameter optimization. 
The number of epochs required for fine-tuning varied for each network as Fig. \ref{fig:train-curves} illustrates. We chose to typically train up to 90-100 epochs and then selected the most appropriate result. In Fig. \ref{fig:hists-acts}, we show the histograms of a selected number of Inception-V3 layers before and after sparsification. Histograms of the sparse model have a greater proportion of zero-values, leading to model acceleration as well as lower entropy, leading to higher activation map compression.

\begin{table}[t]
  \centering
  \resizebox{\columnwidth}{!}
  {  
  \begin{tabular}{c|c|ccccc}
  \toprule
  \textbf{\makecell{Dataset}} & \textbf{Model} &\textbf{Algorithm} & $t=1,000$ & $t=10,000$ & $t=30,000$ &$t=60,000$\\
  \midrule
  \multirow{ 2}{*}{MNIST}& \multirow{ 2}{*}{LeNet-5} & SEG & 1.700$\times$ & 1.700$\times$ & 1.701$\times$ & 1.701$\times$ \\
  \cmidrule(lr{1em}){3-7}
  & & ZVC\cite{rhu2017} & 1.665$\times$ & 1.666$\times$ & 1.666$\times$ & 1.667$\times$\\
  \toprule
  \textbf{Dataset} & \textbf{Model}  & \textbf{Algorithm} & $t=500$ & $t=1,000$ & $t=2,000$ &$t=5,000$\\
   \midrule
   \multirow{ 2}{*}{ImageNet}& \multirow{ 2}{*}{Inception-V3}& SEG & 1.763$\times$ & 1.769$\times$ & 1.774$\times$ & 1.779$\times$\\
   \cmidrule(lr{1em}){3-7}
   & & ZVC\cite{rhu2017} & 1.652$\times$ & 1.655$\times$ & 1.661$\times$ & 1.667$\times$\\
  \bottomrule    
\end{tabular}
  }
\caption{Effect of evaluation set size on compression performance. Varying the size yields minor compression gain changes, indicating that a smaller dataset can serve as a good benchmark for evaluation purposes.}
\label{tab:eval_size} 
\end{table}

  In Fig. \ref{fig:train-curves}, we show how the accuracy and activation map sparsity of MobiletNet-V1 and Inception-V3 changes during training evaluated on the validation sets of CIFAR-10 and ILSVRC2012 respectively. On the same figure, we show the checkpoint selected to report results in Table \ref{tab:spars-results}. Selecting a checkpoint trades-off accuracy with sparsity and which point is chosen depends on the application at hand. In the top row of Fig. \ref{fig:layer-sparsity} we show the percentage of non-zero activations per layer for the five networks. While some layers (e.g. last few layers in MobileNet-V1) exhibit tremendous decrease in non-zero activations, most of the acceleration comes by effectively sparsifying the first few layers of a network. Finally, in Table \ref{tab:comp-spars} we compare our approach to other state-of-the-art methods as well as to a weight pruning algorithm \cite{han2015}. In particular, note that \cite{han2015} is not effective in increasing the sparsity of activations.
  Overall, our approach can achieve both high acceleration (with a slight accuracy drop) and high accuracy (with lower acceleration).

  \begin{table}[t]
    \centering
    \resizebox{\columnwidth}{!}
    {  
    \begin{tabular}{c|c|c|cccc}
    \toprule
    \textbf{Model} & \textbf{Variant} & \textbf{Measurement} & \textbf{float32} & \textbf{uint16} & \textbf{uint12} & \textbf{uint8}\\
    \midrule
    \multirow{4}{*}{\makecell{LeNet-5\\(MNIST)}} & \multirow{2}{*}{Baseline} & Top-1 Acc. & 98.45\% & 98.44\% (-0.01\%) & 98.44\% (-0.01\%) & 98.39\% (-0.06\%) \\    
    & & Compression & - & 3.40$\times$ (1.70$\times$) & 4.40$\times$ (1.64$\times$) & 6.32$\times$ (1.58$\times$) \\    
    \cmidrule(lr{1em}){2-7}
    & \multirow{2}{*}{Sparse} & Top-1 Acc. & 98.48\% (+0.03\%) & 98.48\% (+0.03\%) & 98.49\% (+0.04\%) & 98.46\% (+0.01\%) \\    
    & & Compression & - & 6.76$\times$ (3.38$\times$) & 8.43$\times$ (3.16$\times$) & \textbf{11.16$\times$ (2.79$\times$)} \\     
    \midrule
    \multirow{4}{*}{\makecell{MobiletNet-V1\\(CIFAR-10)}} & \multirow{2}{*}{Baseline} & Top-1 Acc. & 89.17\% & 89.18\% (+0.01\%) & 89.15\% (-0.02\%) & 89.16\% (-0.01\%) \\    
    & & Compression & - & 5.52$\times$ (2.76$\times$)  & 7.09$\times$ (2.66$\times$) & 9.76$\times$ (2.44$\times$)  \\
    \cmidrule(lr{1em}){2-7}
    & \multirow{2}{*}{Sparse} & Top-1 Acc. & 89.71\% (+0.54\%) & 89.72\% (+0.55\%) & 87.72 (+0.55\%) & 89.62\% (+0.45\%) \\    
    & & Compression & - & 5.84$\times$ (2.92$\times$) & 7.33$\times$ (2.79$\times$) & \textbf{10.24$\times$ (2.56$\times$)} \\   
    \bottomrule    
  \end{tabular}
    }
  \caption{Effect of quantization on compression on SEG. LeNet-5 is compressed by 11$\times$ and MobileNet-V1 by 10$\times$. In brackets, we report change in accuracy and compression gain over the float32 baseline.}
  \label{tab:quant_compr_effect} 
  \end{table} 

  \begin{table*}[t]
    \centering
    \scriptsize
    \resizebox{\linewidth}{!}
    {  
    \begin{tabular}{c|cccccccccc}
    \toprule
    \textbf{Dataset} &\textbf{Model} & \textbf{Variant} & \textbf{Bits} &\textbf{Top-1 Acc.}  & \textbf{Top-5 Acc.} & \textbf{SEG} &\textbf{EG \cite{teuhola1978}} & \textbf{HC \cite{han2015}} & \textbf{ZVC \cite{rhu2017}} & \textbf{ZLIB \cite{Gailly1996}} \\
    \midrule
    \multirow{ 3}{*}{MNIST}  & \multirow{ 3}{*}{LeNet-5} & Baseline & float32 & 98.45\% &  - & - & - & - & - & -\\
     & & Baseline &  \multirow{ 2}{*}{uint16} & 98.44\% (-0.01\%) & - & \textbf{3.40$\times$ (1.70$\times$)} & 2.30$\times$ (1.15$\times$) & 2.10$\times$ (1.05$\times$) & 3.34$\times$ (1.67$\times$) & 2.42$\times$ (1.21$\times$)\\
     & & Sparse & & \textbf{98.48\% (+0.03\%)} & - & \textbf{6.76$\times$ (3.38$\times$)} & 4.54$\times$ (2.27$\times$) & 3.76$\times$ (1.88$\times$) & 6.74$\times$ (3.37$\times$) & 3.54$\times$ (1.77$\times$)\\
    \midrule
    \multirow{ 3}{*}{CIFAR-10}  & \multirow{ 3}{*}{MobileNet-V1} & Baseline & float32 & 89.17\% & -& - & - & - & - & -\\
    & & Baseline & \multirow{ 2}{*}{uint16} & 89.18\% (+0.01\%) & - & \textbf{5.52$\times$ (2.76$\times$)} & 3.70$\times$ (1.85$\times$) & 2.90$\times$ (1.45$\times$) & 5.32$\times$ (2.66$\times$) & 3.76$\times$ (1.88$\times$)\\
    & & Sparse &  & \textbf{89.72\% (+0.55\%)} & - & \textbf{5.84$\times$ (2.92$\times$)} & 3.90$\times$ (1.95$\times$) & 3.00$\times$ (1.50$\times$) & 5.58$\times$ (2.79$\times$) & 3.90$\times$ (1.95$\times$)\\
    \midrule
    \multirow{ 13}{*}{ImageNet}  & \multirow{ 4}{*}{Inception-V3} & Baseline & float32 & 75.76\% & 92.74\%& - & - & - & - & -\\
    & & Baseline & \multirow{ 3}{*}{uint16} & 75.75\% (-0.01\%) & 92.74\% (+0.00\%) & \textbf{3.56$\times$ (1.78$\times$)} & 2.42$\times$(1.21$\times$) & 2.66$\times$ (1.33$\times$) & 3.34$\times$ (1.67$\times$) & 2.66$\times$ (1.33$\times$)\\
    & & Sparse &   & \textbf{76.12\% (+0.36\%)} & \textbf{92.83\% (+0.09\%)} & \textbf{5.80$\times$ (2.90$\times$)} & 4.10$\times$ (2.05$\times$) & 4.22$\times$ (2.11$\times$) & 5.02$\times$ (2.51$\times$) & 3.98$\times$ (1.99$\times$)\\
    & & Sparse\_v2 &  & 68.96\% (-6.80\%) & 88.54\% (-4.20\%) & \textbf{6.86$\times$ (3.43$\times$)} & 5.12$\times$ (2.56$\times$) & 5.12$\times$ (2.56$\times$) & 6.36$\times$ (3.18$\times$) & 4.90$\times$ (2.45$\times$)\\
    \cmidrule(lr{1em}){2-11}
      & \multirow{ 4}{*}{ResNet-18} & Baseline & float32 & 69.64\% & 88.99\% & - & - & - & - & - \\
    & & Baseline & \multirow{ 3}{*}{uint16} & 69.64\% (+0.00\%) & 88.99\% (+0.00\%) & \textbf{3.22$\times$ (1.61$\times$)} & 2.32$\times$ (1.16$\times$) & 2.54$\times$ (1.27$\times$) & 3.00$\times$ (1.50$\times$) & 2.32$\times$ (1.16$\times$)\\
    & & Sparse  & & \textbf{69.85\% (+0.21\%)} & \textbf{89.27\% (+0.28\%)} & \textbf{4.00$\times$ (2.00$\times$)} & 2.70$\times$ (1.35$\times$) & 3.04$\times$ (1.52$\times$) & 3.60$\times$ (1.80$\times$) & 2.68$\times$ (1.34$\times$)\\
    & & Sparse\_v2  & & 68.62\% (-1.02\%) & 88.41\% (-0.58\%) & \textbf{5.54$\times$ (2.77$\times$)} & 3.80$\times$ (1.90$\times$) & 4.02$\times$ (2.01$\times$) & 4.94$\times$ (2.47$\times$) & 3.54$\times$ (1.77$\times$)\\
    \cmidrule(lr{1em}){2-11}
      & \multirow{ 4}{*}{ResNet-34} & Baseline & float32 & 73.26\% & 91.43\% & - & - & - & - & -\\
    & & Baseline & \multirow{ 3}{*}{uint16} & 73.27\% (+0.01\%) & 91.43\% (+0.00\%)& \textbf{3.38$\times$ (1.69$\times$)} & 2.38$\times$ (1.19$\times$) & 2.56$\times$ (1.28$\times$) & 3.14$\times$ (1.57$\times$) & 2.46$\times$ (1.23$\times$)\\
    & & Sparse &  & \textbf{73.96\% (+0.70\%)} &  \textbf{91.61\% (+0.18\%)} & \textbf{4.18$\times$ (2.09$\times$)} & 2.84$\times$ (1.42$\times$) & 3.04$\times$ (1.52$\times$) & 3.78$\times$ (1.89$\times$) & 2.84$\times$ (1.42$\times$)\\
    & & Sparse\_v2 &  & 67.74\% (-5.52\%) & 87.90\% (-3.53\%)& \textbf{6.26$\times$ (3.13$\times$)} & 4.38$\times$ (2.19$\times$) & 4.32$\times$ (2.16$\times$) & 5.58$\times$ (2.79$\times$) & 4.02$\times$ (2.01$\times$)\\
    \bottomrule    
  \end{tabular}
    }
  \caption{Compressing activation maps. We report the Top-1/Top-5 accuracy, with the numbers in brackets indicating the change in accuracy. The total compression gain is reported for various state-of-the-art algorithms (in brackets we also report the compression gain without including gains from quantization). SEG outperforms other state-of-the-art algorithms in all models and datasets.}
  \label{tab:compr-results} 
  \end{table*} 

  \textbf{Compression.} We carry our compression experiments on a subset of each network's activation maps, since caching activation maps of the entire training and testing sets is prohibitive due to storage constraints. In Table \ref{tab:eval_size} we show the effect of testing size on the compression performance of two algorithms, SEG and zero-value compression (ZVC) \cite{rhu2017} on MNIST and ImageNet. Evaluating compression yields minor differences as a function of size. Similar observations can be made on all investigated networks and datasets. In subsequent experiments, we use the entire testing set of MNIST to evaluate LeNet-5, the entire testing set of CIFAR-10 to evaluate MobileNet-V1 and 5,000 randomly selected images from the validation set of ILSVRC2012 to evaluate Inception-V3 and ResNet-18/34. The Top-1/Top-5 accuracy, however, is measured on the entire validation sets. 

  \begin{table}[t]
    \centering
    {
    \resizebox{0.9\linewidth}{!}
    {    
    \begin{tabular}{c|cccc}
    \toprule
    \textbf{Dataset} &\textbf{Model} & \textbf{Variant} & \textbf{SEG} &\textbf{EG \cite{teuhola1978}}  \\
    \midrule
    \multirow{ 2}{*}{MNIST}  & \multirow{ 2}{*}{LeNet-5} & Baseline & $k=12$ & $k=9$ \\
     & & Sparse & $k=14$ & $k=0$ \\
    \midrule
    \multirow{ 2}{*}{CIFAR-10}  & \multirow{ 2}{*}{MobileNet-V1} & Baseline & $k=13$ & $k=0$ \\
     & & Sparse & $k=13$ & $k=0$ \\
    \midrule
    \multirow{ 9}{*}{ImageNet}  & \multirow{ 3}{*}{Inception-V3} & Baseline & $k=12$ & $k=7$\\
      &  & Sparse & $k=10$ & $k=0$\\
      &  & Sparse\_v2 & $k=13$ & $k=0$\\  
      \cmidrule(lr{1em}){2-5}
      & \multirow{ 3}{*}{ResNet-18} & Baseline & $k=12$ & $k=10$\\
      &  & Sparse & $k=12$ & $k=0$\\
      &  & Sparse\_v2 & $k=11$ & $k=0$\\      
      \cmidrule(lr{1em}){2-5}
      & \multirow{ 3}{*}{ResNet-34} & Baseline & $k=12$ & $k=9$\\
      &  & Sparse & $k=12$ & $k=0$\\
      &  & Sparse\_v2 & $k=11$ & $k=0$\\     
    \bottomrule    
  \end{tabular}
    }
    }
  \caption{SEG and EG parameter values. Parameters were selected on a separate training set composed of $1,000$ randomly selected images from each dataset. SEG fits the data distribution better by splitting the values into two sets: zero and non-zero. When searching for the optimal parameter value, we fit the parameter to the non-zero value distribution. EG fits the parameter to both sets simultaneously resulting in a sub-optimal solution.}
  \label{tab:compr-params} 
  \end{table}

  In Table \ref{tab:compr-results}, we summarize the compression results. Compression gain is defined as the ratio of activation size before and after compression. We evaluate our method (SEG) against exponential-Golomb (EG) \cite{teuhola1978}, Huffman Coding (HC) used in \cite{han2015}, zero-value compression (ZVC) \cite{rhu2017} and ZLIB \cite{Gailly1996}. We report the Top-1/Top-5 accuracy of the quantized models at $16$ bits. Parameter choice for SEG and EG is provided in Table \ref{tab:compr-params}. SEG outperforms all methods in both the baseline and sparse models validating the claim that it is a very effective encoder for distributions with high sparsity and long tail. \emph{SEG achieves almost $7\times$ compression gain on LeNet-5, almost $6\times$ on MobileNet-V1 and Inception-V3 and more than $4\times$ compression gain in the ResNet architectures while also featuring an increase in accuracy.}
  
  It is also clear that our sparsification step leads to greater compression gains, as can be seen by comparing the baseline and sparse models of each network. When compared to the baseline, the sparse models of LeNet-5, MobileNet-V1, Inception-V3, ResNet-18 and ResNet-34 can be \emph{compressed} by $2.0\times,1.1\times,1.6\times,1.2\times,1.2\times$ respectively \emph{more} than the \emph{compressed} baseline, while also featuring an \emph{increase} in accuracy and an \emph{accelerated} computation of $2.3\times,1.6\times,1.6\times,1.2\times,1.2\times$ respectively. Table \ref{tab:compr-results} also demonstrates that our pipeline can achieve even greater compression gains if slight accuracy drops are acceptable. The sparsification step induces higher sparsity and lower entropy in the distribution of values, which both lead to these additional gains (Fig. \ref{fig:hists-acts}). While all compression algorithms benefit from the sparsification step, SEG is the most effective in exploiting the resulting distribution of values.

  In the bottom row of Fig. \ref{fig:layer-sparsity} we show the compression gain per layer for various networks. Finally, we study the effect of quantization on compression for SEG in Table \ref{tab:quant_compr_effect}. We evaluate the baseline and sparse models of LeNet-5 and MobileNet-V1 at $q=16,12,8$ bits and report compression gain and accuracy. We report the accuracy change and compression gain compared to the floating-point baseline model. \emph{LeNet-5 can be compressed by as much as 11$\times$ with a $0.01\%$ accuracy gain, while MobileNet-V1 can be compressed by $10\times$ with a $0.45\%$ accuracy gain.}

\section{Conclusion}

We have presented a three-stage compression and acceleration pipeline that sparsifies, quantizes and encodes activation maps of CNN's. The sparsification step increases the number of zero values leading to model acceleration on specialized hardware, while the quantization and encoding stages lead to compression by effectively utilizing the lower entropy of the sparser activation maps. The pipeline demonstrates an effective strategy in reducing the computational and memory requirements of modern neural network architectures, taking a step closer to realizing the execution of state-of-the-art neural networks on low-powered devices. At the same time, we have demonstrated that adding a sparsity-inducing prior on the activation maps does not necessarily come at odds with the accuracy of the model, alluding to the well-known fact that sparse activation maps have a strong representational power. Furthermore, we motivated our proposed solution by drawing connections between our approach and sparse coding. Finally, we believe that an optimization scheme in which we jointly sparsify and quantize neural networks can lead to further improvements in acceleration, compression and accuracy of the models.

\noindent
\textbf{Acknowledgments.} We would like to thank Hui Chen, Weiran Deng and Ilia Ovsiannikov for valuable discussions.

{\small
\bibliographystyle{ieee}
\bibliography{../../../mybib}
}

\clearpage
\appendix
\section{Appendix overview}

In the appendices, we discuss (1) the parameter choice of the regularization parameter $\alpha_l$ for sparsification for each layer $l$ of a network (Sec. \ref{sec:regparam})
and (2) present the exponential-Golomb algorithm for completeness (Sec. \ref{sec:exp-golomb}).

\section{Regularization parameter} 
\label{sec:regparam}
We present the regularization parameters per layer for each network in Tables \ref{tab:reg-lenet5}, \ref{tab:mobilenetv1}, \ref{tab:reg-imagenet}. While we finely-tuned the regularization parameter of each layer for the smaller networks (LeNet-5, MobileNet-V1), we chose a global regularization parameter for the bigger ones (Inception-V3, ResNet-18, ResNet-34). Finely-tuning the regularization of each layer can yield further sparsity gains, but also adds an intensive search over the parameter space. Instead, for the bigger networks, we showed that even one global regularization parameter per network is sufficient to successfully sparsify a model. For layers not shown in the tables, the regularization parameter was set to $0$. Note that we never regularize the final layer $L$ of each network.

\begin{table}[h]
  \centering
  \begin{tabular}{cccc}
  \toprule
  \textbf{LeNet-5} & conv1 & conv2 & fc1 \\
  \midrule
  $\alpha_l$ & $0.25\times10^{-5}$ & $2.00\times10^{-5}$ & $5.00\times10^{-5}$ \\
  \bottomrule    
\end{tabular}
\caption{LeNet-5 \cite{lecun98} regularization parameters.}
\label{tab:reg-lenet5} 
\end{table} 

\begin{table}[h]
  \footnotesize
  \centering
  \resizebox{\linewidth}{!}
  {  
  \begin{tabular}{ccccccccccc}
  \toprule
  \textbf{MobileNet-V1} & Conv/s2 & \makecell{Conv dw/s1 \\ Conv / s1} & \makecell{Conv dw/s2 \\ Conv / s1} & \makecell{Conv dw/s1 \\ Conv / s1} & \makecell{Conv dw/s2 \\ Conv / s1} & \makecell{Conv dw/s1 \\ Conv / s1} & \makecell{Conv dw/s2 \\ Conv / s1} & 5$\times$
  \makecell{Conv dw / s1 \\
  Conv / s1} & \makecell{Conv dw/s2 \\ Conv / s1} & \makecell{Conv dw/s2 \\ Conv / s1}\\
  \midrule
  $\alpha_l$ & $15\times10^{-8}$ & $15\times10^{-8}$ & $15\times10^{-8}$ & $15\times10^{-8}$ & $1\times10^{-8}$ & $1\times10^{-8}$ & $1\times10^{-8}$ & $1\times10^{-8}$ & $2\times10^{-8}$ & $2\times10^{-8}$ \\
  \bottomrule    
\end{tabular}
  }
\caption{MobileNet-V1 \cite{howard2017} regularization parameters.}
\label{tab:mobilenetv1}
\end{table}

\begin{table}[h]
  \centering
  \begin{tabular}{c|cc}
  \toprule
  \textbf{Model} & \textbf{Variant} & \makecell{$\alpha_l$ \\$(l=1,\hdots,L-1)$}\\
  \midrule
  \multirow{ 2}{*}{Inception-V3} & Sparse & 1$\times10^{-8}$\\
  \cmidrule(lr{1em}){2-3}
  & Sparse\_v2 & 1$\times10^{-7}$ \\
  \midrule
  \multirow{ 2}{*}{ResNet-18} & Sparse & 1$\times10^{-8}$\\
  \cmidrule(lr{1em}){2-3}
  & Sparse\_v2 & 1$\times10^{-7}$ \\  
  \midrule
  \multirow{ 2}{*}{ResNet-34} & Sparse & 1$\times10^{-8}$\\
  \cmidrule(lr{1em}){2-3}
  & Sparse\_v2 & 5$\times10^{-8}$ \\    
\bottomrule    
\end{tabular}
\caption{Inception-V3 \cite{szegedy2016} and the ResNet-18/34 \cite{he2016} regularization parameters.}
\label{tab:reg-imagenet} 
\end{table} 

\section{Exponential-Golomb} 
\label{sec:exp-golomb}
We provide pseudo-code for the encoding and decoding algorithms of exponential-Golomb \cite{teuhola1978} in Alg. \ref{alg:exp_golomb}.

\begin{algorithm}[h]  
  \caption{Exponential-Golomb}
  \begin{algorithmic}  
  \STATE \textbf{Input}: Non-negative integer $x$, Order $k$\\  
  \textbf{Output}: Bitstream $y$\\  
  $\text{function \textbf{encode\_exp\_Golomb} }( x , k )$\\
  \{\\
    \INDSTATE If $k == 0$:\\
    \INDSTATE[2] $y = \text{encode\_exp\_Golomb\_0\_order} ( x )$\\    
    \INDSTATE Else:\\
    \INDSTATE[2] $q = \text{floor} ( x / 2^k )$\\
    \INDSTATE[2] $q_c = \text{encode\_exp\_Golomb\_0\_order} ( q )$\\    
    \INDSTATE[2] $r = x \mod 2^k$\\
    \INDSTATE[2] $r_c = \text{to\_binary} ( r , k )$ // $\text{to\_binary} ( r , k )$ converts $r$ into binary using $k$ bits.\\
    \INDSTATE[2] $y = \text{concatenate} (q_c , r_c)$\\
    \INDSTATE Return $y$\\   
  \}\\
  \STATE \textbf{Input}: Bitstream $x$, Order $k$\\  
  \STATE \textbf{Output}: Non-negative integer $y$ \\
  $\text{function \textbf{decode\_exp\_Golomb} }( x )$\\
  \{\\
  \INDSTATE If $k == 0$:\\
  \INDSTATE[2] $y , l = \text{decode\_exp\_Golomb\_0\_order} ( x )$\\    
  \INDSTATE Else:\\
  \INDSTATE[2] $q , l = \text{decode\_exp\_Golomb\_0\_order} ( x )$\\
  \INDSTATE[2] $r = \text{int} ( x [ l : l + k ] )$
  \INDSTATE[2] $y = q \times ( 2^k ) + r$
  \INDSTATE Return $y$\\   
  \}\\  
  \STATE \textbf{Input}: Non-negative integer $x$\\  
  \textbf{Output}: Bitstream $y$\\    
  $\text{function \textbf{encode\_exp\_Golomb\_0\_order} }( x )$\\
  \{\\
  \INDSTATE $q = \text{to\_binary} ( x + 1 )$\\
  \INDSTATE $q_\text{len} = \text{length} ( q ) $ \\
  \INDSTATE $p = ``0" * (q_{\text{len}}-1)$ // \text{replicates ``0'' $q_{\text{len}}-1$ times.} \\
  \INDSTATE $y = \text{concatenate} ( p , q )$\\
  \INDSTATE Return $y$\\
  \}\\
  \STATE \textbf{Input}: Bitstream $x$\\  
  \STATE \textbf{Output}: Non-negative integer $y$, Non-negative integer $l$ \\
  $\text{function \textbf{decode\_exp\_Golomb\_0\_order} }( x )$\\
  \{\\
  \INDSTATE $p = \text{count\_consecutive\_zeros\_from\_start} ( x )$ // \text{consecutive zeros of $x$ before the first ``1''.}\\
  \INDSTATE $y = \text{int} ( x [p:2\times p + 1] ) - 1$\\
  \INDSTATE $l = 2 \times p + 1$\\
  \INDSTATE Return $y,l$\\
  \}\\  
  //The notation $x[a:b]$, follows the Python rules, i.e. selects characters in the range $[a,b)$\\
  \end{algorithmic}
  \label{alg:exp_golomb}
 \end{algorithm}

\end{document}